\documentclass{article}
\pdfoutput=1
\usepackage{amssymb,amsmath}
\usepackage{algorithm}
\usepackage{algorithmic}
\usepackage{enumitem}
\usepackage{graphicx}
\graphicspath{{./arXivSubmission/}}
\usepackage{booktabs}
\usepackage[table]{xcolor}
\usepackage{epsfig, subfigure}
\usepackage{url}
\usepackage{color}   
\usepackage{hyperref}
\usepackage{bbm}
\usepackage{authblk}
\usepackage{algorithm}
\usepackage{float}
\usepackage{wrapfig}

\newcommand {\R}{\mathbb {R}}
\newcommand {\reals}{\mathbb {R}}
\newcommand {\x}{\mathbf {x}}
\newcommand {\bs}{\mathbf{s}}

\newcommand {\bv}{\mathbf{v}}
\newcommand {\bx}{\mathbf{x}}
\newcommand {\bc}{\mathbf{c}}
\newcommand {\bq}{\mathbf{q}}
\newcommand {\bfe}{\mathbf{e}}

\newcommand {\cN}{\mathcal {N}}

\newcommand {\B}{\mathbf {B}}

\newcommand{\be}{\begin{equation*}}
\newcommand{\ee}{\end{equation*}}

\DeclareMathOperator{\diag}{diag}

\title{High Dimensional Data Modeling Techniques for Detection of Chemical Plumes and Anomalies in Hyperspectral Images and Movies}
\author{Yi (Grace) Wang\thanks{Y.~Wang is with the Mathematics Department, Syracuse University, Syracuse, NY 13244, United States. },
Guangliang Chen\thanks{G.~Chen is with Department of Mathematics and Statistics, San Jos\'e State University, San Jos\'e, CA 95192, United States.}, and Mauro Maggioni\thanks{M.~Maggioni is with the Departments of Mathematics, Electrical and Computer Engineering, Computer Science, Duke University, Durham, NC 27708, United States.}
\thanks{The work of M.~Maggioni was supported by the ATD program funded by NSF, NGA and DTRA, under award \# 1222567, and by the DHS CAXSI program.}}
\date{}

\begin{document}
\maketitle
\begin{abstract}
We briefly review recent progress in techniques for modeling and analyzing hyperspectral images and movies, in particular for detecting plumes of both known and unknown chemicals. For detecting chemicals of known spectrum, we extend the technique of using a single subspace for modeling the background to a ``mixture of subspaces" model to tackle more complicated background. Furthermore, we use partial least squares regression on a resampled training set to boost performance. For the detection of unknown chemicals we view the problem as an anomaly detection problem, and use novel estimators with low-sampled complexity for intrinsically low-dimensional data in high-dimensions that enable us to model the ``normal'' spectra and detect anomalies. We apply these algorithms to benchmark data sets made available by the Automated Target Detection program co-funded by NSF, DTRA and NGA, and compare, when applicable, to current state-of-the-art algorithms, with favorable results.
\newline
\smallskip

\end{abstract}


\section{Introduction}
The ability to remotely sense and analyze chemical plumes has become increasingly important in many civilian and military applications. Hyperspectral imaging sensors that operate in the
long-wave infrared (LWIR) part of the spectrum are particularly well
suited for these chemical-sensing tasks, because their wide fields of view and high spectral resolutions allow many square kilometers to be imaged almost simultaneously and even optically thin chemical clouds to be detected.
Data collected by these sensors has the form of $m \times n \times p$ arrays, where $m, n$ are the spatial dimensions and $p$ is the spectral dimension. Under physically reasonable assumptions and simplifications 
\cite{Manolakis-review-IRLW},
the following linear mixing model is obtained:
\begin{equation} \label{eq:linear_model}
\x = \sum_{1\le i \le N} g_i \bs_i + \bv,
\end{equation}
where $\x \in \R^p$ represents the radiance spectrum in the scene, $N$ the number of chemicals of interest, $\bs_i\in\R^p$ the signature spectrum for the $i$-th chemical of interest, $g_i\ge 0$  the amount of chemical $i$, and $\bv$ the radiance spectrum of the background. In practice, it is usually assumed that $N \leq 3$. In this paper, due to the data under consideration, we consider the case $N=1$ (thus, $\x =  g \bs  + \bv$). However, the methods discussed in this paper can be naturally extended to the cases where $N>1$.

To effectively separate the chemical clouds from the background clutter, one needs to choose a proper model for the background radiation $\bv$. Current approaches (e.g.~\cite{ACE, Manolakis-subspace}) often represent the background by a single Gaussian distribution or subspace and then derive corresponding statistical estimators which assign detection scores to each spectrum (see the review \cite{Manolakis-review-IRLW}). These detection algorithms have shown their effectiveness in many applications, but there is room for improvement. In particular, background often consists of heterogeneous regions (such as sky, mountain, desert; see e.g.~\cite[Fig.~9]{Manolakis-review-IRLW}) which may require separate subspaces to better capture the complexity of the background. Some other methods tackle the problem alternatively~\cite{jointSparse,EnsembleLearning,nonGaussian,Bertozzi}, including the case when the target plume is unknown~\cite{Theiler}.

Our contributions in this work are threefold: (a) we extend the single-subspace model~\cite{Manolakis-subspace} for modeling the background to a ``mixture of subspaces'' model; (b) we propose techniques to enhance the detectability of the chemical plume region, moving beyond classical least squares and using resampling techniques to boost detection; and (c) when detection of chemicals with unknown signature is of interest, we propose a flexible anomaly detection procedure which efficiently constructs an empirical model of ``normal'' spectra and flags anomalous ones according to likelihood assigned by the empirical model. In particular, (c) is achieved by applying an efficient multiscale transform~\cite{GMRA_ACHA12} to the hyperspectral spectra of a training frame and then model the density of the transformed pixels for computing the likelihood of each spectrum in any testing frame: if the testing frame contains gas, then the gas region will be flagged as an anomaly if it is assigned low likelihood values.

We consider two different scenarios, depending on whether we are given only a single hyperspectral cube or a time series of them. In Scenario (I), we use the available cube both for learning an empirical model for the background and for chemical detection, while in Scenario (II) we use the first few frames (assumed to be without chemical plume) for background modeling and any subsequent frame for testing for anomalies based on the learned background model. If the sensor is assumed stationary, one could use the temporal variation of the spectrum at each location to detect changes and detect anomalies: we do {\em{not}} assume that the sensor is stationary (albeit it is such in most of the data we consider) and therefore we do not use spatial coherence information across the images. This makes the problem harder, but the proposed solution applicable in a wider variety of settings.

\section{Previous Work for Known Target Signature\label{sec:prev}}
Based on different assumptions on the background radiation $\bv$, different detection algorithms have been proposed for the setting when the signature of the target plume is known. In this section we review some of the existing approaches following the presentation of \cite{Manolakis-review-IRLW}.

\subsection{Gaussian Models}
The simplest and most practical algorithm for chemical gas detection uses the linear model in \eqref{eq:linear_model} and assumes normally distributed background clutter with known covariance, that is,
\be
\x = \bs g+ \bv, \qquad  \bv \sim \cN(\mu,\sigma^2\Sigma)
\ee
where the plume spectral signature $\bs$ is also assumed to be  known. The coefficients $g$ and $\sigma^2$ can be estimated by maximizing the likelihood from a given sample. The gas detection problem can be formalized as that of testing the hypotheses
\begin{eqnarray}
H_0 \,\,:\,\,  \text{Plume absent } (g  = 0); \quad
H_1 \,\,:\,\, \text{Plume present } (g > 0). \nonumber
\end{eqnarray}
By applying the generalized likelihood ratio test (GLRT) one obtains the following detector:
\begin{equation}
T_{\text{NMF}}(\bx \mid \mu, \Sigma,\bs ) = \frac{(\bs^T\Sigma^{-1}\tilde\bx)^2}{(\bs^T \Sigma^{-1}\bs)(\tilde\bx^T \Sigma^{-1}\tilde\bx)}, \quad \tilde\bx = \bx - \mu
\label{eq:NMF}
\end{equation}
This is known as the normalized matched filter (NMF) and also the adaptive cosine or coherence estimator (ACE)~\cite{ACE}. Since the NMF requires estimation and inversion of the background covariance matrix $\Sigma$, they are approximated in the following way in order to avoid numerical instability and to obtain robust detectors:
\be
\hat{\Sigma} = \sum_{k=1}^p \lambda_k \bq_k \bq_k^T + \delta I, \qquad  \hat{\Sigma}^{-1} = \sum_{k=1}^p \frac{1}{\lambda_k+\delta} \bq_k \bq_k^T \nonumber
\ee
where $\lambda_k$ and $\bq_k$ ($k=1,\cdots,p$) are the eigenvalues and eigenvectors of the sample variance, $\delta$ is a regularization parameter and can be chosen as certain percentile of the values of $\{\lambda_k\}$. For the data of our interest in this paper, the performance of ACE is not sensitive to the choice of $\delta$ as long as it is moderate, say between 30 and 80 percentiles of $\{\lambda_k\}$. In the experiments of this paper, we choose $\delta$ as the median of $\{\lambda_k\}$.

\subsection{Subspace Models}
Another category of approaches  for chemical plume detection assumes that the background clutter can be well represented by a low-dimensional subspace. In that case, the signal model becomes
\be
\bx = \bs g + \mu + B \bc + \epsilon, \quad \epsilon \sim \cN(0,\sigma^2 I)
\label{eq:ssm}
\ee
where $\mu, B$ represent a fixed point and a basis for the background subspace and $\epsilon$ is the additive noise. To find $\mu, B$, principal component analysis (PCA) is applied to the background spectra.  Similarly to before, the quantities $g$, $\bc$ and $\sigma$ are estimated by maximizing the likelihood. Due to the assumption of normally distributed error $\epsilon$, the maximum likelihood estimators (MLE) of $g$ and $\bc$ may be computed by least squares.

The GLRT approach yields the following detector~\cite{Manolakis-subspace}:
\begin{equation}
T_{\text{NSS}}(\bx \mid \mu, \B,\bs) = \frac{\| P_b^{\perp} \bx\|^2}{\| P_{tb}^{\perp} \bx\|^2}
\label{eq:NSS}
\end{equation}
where $P_b^{\perp}$ and $P_{tb}^{\perp}$ are projection matrices:
\begin{eqnarray}
P_b^{\perp} &=& I - B(B^TB)^{-1}B^T; \nonumber \\
P_{tb}^{\perp}&=& I - A(A^TA)^{-1}A^T, \;\; A=[\bs \, B].\nonumber
\end{eqnarray}
Note that $\| P_b^{\perp} \bx\|$ and $\| P_{tb}^{\perp} \bx\|$ are respectively the orthogonal distances from the background and target-background subspaces.

\subsection{Multiple Plumes\label{sec:multiPlume}}
These algorithms can be naturally extended to the case where $N>1$, i.e., there are more than one chemical plumes. The corresponding detectors can be written as:
    $$
    T_{\text{NMF}}(\bx \mid \mu, \Sigma,S ) = \frac{(\tilde\bx^T\Sigma^{-1}S)(S^T \Sigma^{-1}S)^{-1}(S^T\Sigma^{-1}\tilde\bx)}{\tilde\bx^T \Sigma^{-1}\tilde\bx}, \quad \tilde\bx = \bx - \mu
    $$
    where $S=[s_1,\cdots, s_N]$ and $s_i \in \reals^p$ is the signature spectrum for the $i^{\text{th}}$ chemical plume, and
$$
T_{\text{NSS}}(\bx \mid \mu, \B,S) =  \frac{\| P_b^{\perp} \bx\|^2}{\| P_{tb}^{\perp} \bx\|^2}
$$
where $P_b^{\perp}$ and $P_{tb}^{\perp}$ are projection matrices:
\begin{eqnarray}
&&P_b^{\perp} = I - B(B^TB)^{-1}B^T; \nonumber \\
&&P_{tb}^{\perp}= I - A(A^TA)^{-1}A^T, \;\; A=[S \, B].\nonumber
\end{eqnarray}

\section{Mixture models and Enhancement with Known Target Signature\label{sec:alg}}
\label{sec:mixmodels}

In this section, we describe both the mixture models and some techniques to enhance the detection. Our algorithm is presented at the end of this section.

\subsection{Mixture Models}
Gaussian mixture models $\bv \sim \sum_j \pi_j \cN(\mu_j,\Sigma_j)$, where $\cN$ represents the Normal distribution, may be used to capture complex backgrounds. We are particularly interested in the case where $\Sigma_j$ is rank-deficient, and therefore $\cN(\mu_j,\Sigma_j)$ is supported on an affine subspace spanned by $\mathrm{cols}(B_j)$, the columns of a matrix $B_j$. More generally, in a {\em{subspace mixture model}} $\bv\sim\sum_j\pi_j\mathcal{S}(\mu_j,B_j)$ where $\mathcal{S}(\mu_j, B_j)$ is a probability measure with mean $\mu_j$ and support contained in the subspace spanned by the columns of $B_j$.

 The model parameters $\pi_j$ and $\Theta_j$ ($(\mu_j,\Sigma_j)$ or $(\mu_j,B_j)$) in each case %
can be estimated by $K$-means, $K$-means-like subspace clustering algorithms (e.g.~\cite{SCC,LBF,CM:CVPR2011}), fast multiscale techniques \cite{vcip}, or Expectation-Maximization (EM) methods,
through iterations between updating cluster assignments and model parameters.

The signal $\x$ is then assigned to the cluster that maximizes the estimation of $\pi_j p(\x|\Theta_j)$:
\[\hat j = \arg\max_j \pi_j p(\bx \mid \Theta_j).\]

Given a target plume signature $\bs$, the mixture versions of the two estimators, NMF (also known as ACE) and NSS, are given by
\begin{equation}
\begin{aligned}
&T_{\text{mixNMF}}(\bx \mid \bs, \{\pi_j,\,\Theta_j\}) =
T_{\text{NMF}}(\bx \mid \bs, \mu_{\hat j},\Sigma_{\hat j});\\
&T_{\text{mixNSS}}(\bx \mid \bs, \{\pi_j,\,\Theta_j\}) =T_{\text{NSS}}(\bx \mid \bs, \mu_{\hat j},B_{\hat j}).
\label{eq:dect}
\end{aligned}
\end{equation}

Alternatively, for the subspace mixture model, we may simply use the coefficient $g$ as the detector and solve for it by least squares. Specifically, letting $\hat j$ be defined as above, it is easy to show that the least squares estimator $\hat g$ of $g$ is the first entry of
$$\hat{\beta} = (A_{\hat j}^TA_{\hat j})^{-1}A_{\hat j}^T(\bx - \mu_{\hat j}), \quad A_{\hat j} = [\bs\, B_{\hat j}].$$
This yields the mixture Linear Coefficient (LC) estimator
$$
T_{\text{mixLC}}(\bx \mid \bs, \{\pi_j,\, \Theta_j\}) = \max\{\hat{g}, 0\}.
$$

The knowledge about the background complexity can be used to choose the appropriate number of Gaussians (or subspaces), e.g., for the data in Section~\ref{sec:MITdata}, we use three components since the scene has sky, mountain and ground areas. In practice, the detection performance is not sensitive when this number is overestimated and there are enough sample spectra. These mixture models can also be applied with the detectors that handle multiple plumes.

Finally, in section \ref{sec:anomalydetection} we consider a related but different family of multiscale models for background spectra, in the context of anomaly detection in hyperspectral movies, in the setting when the spectral signature $\mathbf{s}$ of the chemical of interest is not given, nor the frames in the movies are assumed to be registered.

\subsection{Enhancement Techniques}
We propose a few enhancement techniques for background estimation to reduce the affects of the contamination:

\subsubsection{Outlier Removal}

We detect and remove a small fraction of the spectra that are dissimilar to the main part of the data, in terms of magnitude or connectivity (in this paper we simply compute the sum of squares of each spectrum as its magnitude and classify those spectra with largest magnitudes as outliers).

\subsubsection{Resampling Enhancement}
This technique is relevant only when we are in Scenario (I). For this goal we utilize an iterative scheme. We first choose a few likely background spectra based on a reliable detection score (output of a detection algorithm, e.g.~ACE), and then select their spatial neighbors as well, since adjacent pixels are very likely to be of the same category. Using these selected spectra, the background model parameters are re-estimated and the detection statistics are re-deduced accordingly.
\begin{algorithm}[htbp!]                      
\caption{Resampling Enhancement}          
\label{alg:rs}                           
\begin{algorithmic}[1]                    

    \REQUIRE $\{\bx_i\}_{i=1}^{mn} \subset \reals^p$: spectra; $\{T_i\}_{i=1}^{mn} \subset \reals$: detection score; $\bs \in \reals^p$: plume signature; $\tau_1 \in (0,1)$: portion.
    \ENSURE $\{y_i\}_{i=1}^{mn} \subset \reals$: enhanced detection score.
    \STATE Sort $T_i$ s.t., $T_{(1)} \leq \cdots T_{(mn)}$.
    \STATE Choose $\delta_1 = T_{([\tau_1mn])}$. Let $A_1:=\{\bx_i:\;T_i \leq \delta_1 \}$.
    \STATE Let $B$ be the union of $A_1$ and the 4 (above, below, left and right) spatial neighbors of $x_i \in A$.
    \STATE Re-estimate model parameters $\Theta_j$'s from $B$.
    \STATE Compute the statistic (detection score) based on the updated model.
\end{algorithmic}
\end{algorithm}
\subsubsection{Partial Least Squares Regression (PLSR) Enhancement}
PLSR~\cite{plsr} reduces the dimensionality of the data in the way that the covariance between predictors and responses is maximized. The response is re-estimated on the reduced predictors. In our problem, PLSR is applied so that the radiance data is projected to a subspace that is most relevant to the detection score (output of a detection algorithm, or enhanced detection score resulted from resampling). A new and enhanced detection score is computed from the projected radiance data. In fact, we select a certain amount of spectra which are most likely to be background as well as chemical clouds and apply PLSR on the selections.

\begin{algorithm}[htbp!]                      
\caption{PLSR Enhancement}          
\label{alg:plsr}                           
\begin{algorithmic}[1]                    

    \REQUIRE $\{\bx_i\}_{i=1}^{mn} \subset \reals^p$: spectra; $\{T_i\}_{i=1}^{mn} \subset \reals$: detection score; $\bs \in \reals^p$: plume signature; $\tau_2,\,\tau_3 \in (0,1)$: portions.
    \ENSURE $\{y_i\}_{i=1}^{mn} \subset \reals$: enhanced detection score.
    \STATE Sort $T_i$ s.t., $T_{(1)} \leq \cdots T_{(mn)}$.
    \STATE Choose $\delta_2 = T_{([\tau_2mn])}$ and $\delta_3 = T_{([(1-\tau_3)mn])}$. Let $A_2:=\{\bx_i:\;T_i \leq \delta_2 \}$ and $A_3:=\{\bx_i:\;T_i \geq \delta_3 \}$.
    \STATE Apply PLSR on the selected pairs $(\bx_i,T_i)$ with $\bx_i \in A_2 \bigcup A_3$ to obtain parameters $\beta \in \reals^p$ and $\beta_0 \in \reals$.
    \STATE $y_i = \beta^T \bx_i + \beta_0$, $i = 1,\cdots,mn$.

\end{algorithmic}
\end{algorithm}

\subsection{Algorithm for Plume Detection in Hyperspectral Images or Movies}
We propose a unifying algorithm (see Algorithm.~\ref{alg:cpd}) that can detect chemical plumes in both scenarios: in Scenario (I) when we have only a single hyperspectral cube, we incorporate the enhancement techniques proposed in the previous section to learn the background via mixture modeling;
in Scenario (II) where we are given a time series of hyperspectral cubes, we assume that the first few frames were collected before the chemical release so we may use their spectra for background modeling. Afterwards, we select a detector (from $T_{\text{mixNMF}}, T_{\text{mixNSS}}, T_{\text{mixLC}}$)) and apply it to the given cube(s).

\begin{algorithm}[htbp!]                     
\caption{Chemical plume detection through mixture background modeling in hyperspectral images or movies }          
\label{alg:cpd}                       
\begin{algorithmic}[1]                    

    \REQUIRE $\{\bx_i^{(\ell)}\}_{i=1}^{mn}, 1\le \ell \le L$: $L$ hyperspectral frame(s); $\bs \in \reals^p$: plume signature, and detector (one of the $T_{\text{mixNMF}}, T_{\text{mixNSS}}, T_{\text{mixLC}}$).
    \ENSURE $\{T(\bx_i^{(\ell)})\} \subset \reals$: detection score
    \STATE Fit a mixture model $\{\Theta_j\}$ to the given cube (when $L=1$) or the first few clean frames (when $L>1$). Apply the enhancement techniques if relevant.
    \STATE \textbf{for} each frame $\ell = 1, \ldots, L$,

    \indent (1) Assign spectra of the $\ell$-th frame to nearest component models by maximizing the estimator of $\pi_j p(\x|\Theta_j)$ \\
    \indent (2) Evaluate the given detector for all spectra of the frame (within the corresponding clusters) to produce detection scores

    \textbf{end for}
\end{algorithmic}
\end{algorithm}

\section{Anomaly Detection in hyperspectral movies with Unknown Target Signature}
\label{sec:anomalydetection}

In this section we consider the problem of anomaly detection instead of detection of known chemicals.  In particular, we are interested in the application to hyperspectral movies, where at unknown time and location a gas is released and the gas plume needs to be tracked.
We think of the chemical plume region, once released, as a set of anomalous spectra, when compared against the background clutter, and thus we base detection on accurately estimating the probability measure modeling the space of the background spectra and computing the likelihood scores of every spectrum in any testing frame relative to the density model.

In the original paper on Geometric Multi-Resolution Analysis (GMRA) \cite{GMRA_ACHA12}, an approach for approximating probability distributions in high-dimensions using the intrinsically low-dimensional GMRA structure was suggested, and those ideas were further developed in \cite{vcip,MM_GeometricEstimationAsilomar}.
In this case we are not given signatures of spectra to detect; instead we are given one or more hyperspectral scenes defined ``normal'' (a training set), and given a new hyperspectral scene we are interested in deciding if its spectra are normal or present ``anomalies''. We model this problem as follows: we assume that there is an unknown probability measure $\nu$ in $\mathbb{R}^p$ from which ``normal'' spectra are drawn. The training set $X:=\{\bx_i\}_{i=1}^N\subseteq\mathbb{R}^p$ consisting of all the spectra in the training hyperspectral scenes is modeled as $n$ i.i.d. samples from $\nu$.\footnote{Clearly, independence is a rather strong assumption, but could be relaxed with only rather minor technical difficulties to more general settings that accommodate mild dependencies.}
We use these $n$ samples to learn a probability measure $\hat\nu_{X}$ approximating, in a suitable sense, $\nu$. Given a new scene, i.e. a new set of samples $X^{\text{new}}=\{\bx^{\text{new}}_i\}$, we could ask what is the probability of seeing $\bx^{\text{new}}_i$ if it was sampled from $\nu$, and call $\bx^{\text{new}}_i$ an anomaly if this probability is below a certain threshold. Unfortunately this does not make sense since typically $\nu$ (and our estimator $\hat\nu_X$) do not assign positive probability to any point. Often one then replaces probabilities by probability densities and associated likelihoods. An alternative is to replace the question above by evaluating the probability (according to $\hat\nu_X$) of seeing a point within distance $r$ from $\bx^{\text{new}}_i$, and decide whether $\bx^{\text{new}}$ is an anomaly or not based on a threshold on such probability, i.e. we declare $\bx^{\text{new}}_i$ an anomaly if $\mathbb{P}_{\hat\nu_X}(B_r(\bx^{\text{new}}_i))<\eta$. The values of $\eta$ and $r$ tune the sensitivity of the anomaly detection. We may choose them by first fixing $\eta$, then choosing $r$ minimizing type I or type II error (or other similar criterion), or by choosing $r$ to be smallest value such that $\mathbb{P}_{\hat\nu_X}(B_r(\bx^{\text{val}}_i))>\eta$ for all $i$'s, where $\{\bx^{\text{val}}_i\}$ is a validation data set (possibly extracted and excluded from a training set). Then as we vary $\eta$ we obtain a ROC for the anomaly detector (assuming we know the ground truth).

The key problem here is to efficiently construct an estimator $\hat\nu_{X}$: this is a challenging task, since $\nu$ is in $\mathbb{R}^p$, with $p$ typically large ($p>100$ is common). This is accomplished by using techniques based on GMRA, originally suggested in \cite{GMRA_ACHA12}, and further developed and analyzed in \cite{vcip,MMS:NoisyDictionaryLearning}. We have no space to describe the details of these constructions. At a high level, one can think of GMRA as a principled and efficient way of encoding a data set that, while high-dimensional, is intrinsically low-dimensional, at different levels of accuracy, by first constructing a multiscale tree decomposition of the data, and then in each node of the tree, corresponding to a portion of the data, construct a data-driven low-dimensional projection of the data, and further compressing these projections by encoding the difference between the projection at one node and those of children nodes in the tree \cite{GMRA_ACHA12}. The results in \cite{MMS:NoisyDictionaryLearning} guarantee that under suitable geometric assumptions about the data -- essentially the data should be close to an (unknown) low-dimensional manifold -- GMRA run on the data will efficiently construct sparse representations of the data. In order to extend this construction to the estimation of probability measure, at each node of the GMRA tree we estimate a simple probability measure (e.g. a low-rank Gaussian) and combine these measures appropriately to construct a probability measure $\hat\nu_X$ approximating $\nu$. The results in \cite{vcip,MM_GeometricEstimationAsilomar} guarantee that -- under suitable geometric assumptions on the data and on the regularity of $\nu_X$ -- the estimator $\hat\nu_{X}$ gets increasingly closer to $\nu$ (in Wasserstein distance) with high probability as $N$ increases, and with a rate that depends only on the intrinsic dimension $d$ of the data and not on the ambient dimension $p$. Furthermore, these constructions are yielded by efficient algorithms, having complexity essentially linear in the training data size $pN$, with a constant depending essentially on the intrinsic dimension of the data $d$. In the setting of hyperspectral images considered here, $N=mn$ is the number of pixels in an image. The intrinsic dimension of the data, measured by Multiscale SVD \cite{MM:MultiscaleDimensionalityEstimationAAAI,LMR:MGM1} is often very small in hyperspectral data, between $1$ and $5$. Moreover, these algorithms \cite{vcip}-\cite{MM_GeometricEstimationAsilomar} easily allow to quickly (in $O_d(\log N)$) incorporate new samples in an online fashion, in both the GMRA construction and the estimator $\hat\nu_X$. Finally, the underlying GMRA construction, simultaneously to the above, yields a dictionary that sparsifies the data (in our case the spectra) \cite{MMS:NoisyDictionaryLearning,LMR:MGM1}, enabling the compression of the data. These representations also lead to a variation of Compressed Sensing where the data model is that of a nonlinear manifold, with extremely efficient inversion algorithms that do not require the solution of high-dimensional convex optimization problems \cite{vcip}.

\begin{algorithm}[htbp!]
\caption{Multiscale-transform based Density Estimation}
\label{alg:density_est}
\begin{algorithmic}[1]

\REQUIRE Data set $X_n$
\ENSURE Multiscale densities $\{\hat\nu_{j,k}\}_{j\ge 0, k\in\Lambda_j}$\\
\STATE Apply GMRA to the training data to obtain a multiscale dictionary
$\{c_{j,k}, \Phi_{j,k}, w_{j,k}, \Psi_{j,k}\}_{j>0, k\in \Lambda_j}$
\STATE For each $j>0, k\in \Lambda_j$, apply the Geometric Wavelet Transform to
the data in each $C_{j,k}$ and obtain low dimensional coefficients
\STATE Apply a density estimator (e.g.~KDE~\cite{KDE-IM-http}) to each set of geometric scaling and
wavelet coefficients corresponding to each $C_{j,k}$, and obtain density estimates
$\hat\nu_{j,k}$; also record $\hat\pi_{j,k}$, the empirical probability of a point belonging to $C_{j,k}$.

\STATE By testing on a validation set, select an optimal scale $j^*$ and return the mixture model $\sum_{k} \hat\pi_{j^*,k}\hat\nu_{j^*,k}$.
\end{algorithmic}
\end{algorithm}

\section{Experimental Results}
In this section we consider both synthetic data and some real data sets in the Colorado State Repository (will write CSR for short thereafter) made available through the ATD program ({\em Algorithms for Threat Detection}), co-sponsored by NSF, DTRA and NGA. We start with synthetic data to present the functionality of our models in simple situations, and subsequently consider real data sets, demonstrating the effectiveness of our methods in applications.

\subsection{Synthetic Data Sets}

\subsubsection{Gaussian Mixture Models\label{sec:GaussData}}
\begin{figure}[htbp!]
\centering
\includegraphics[width=0.45\textwidth]{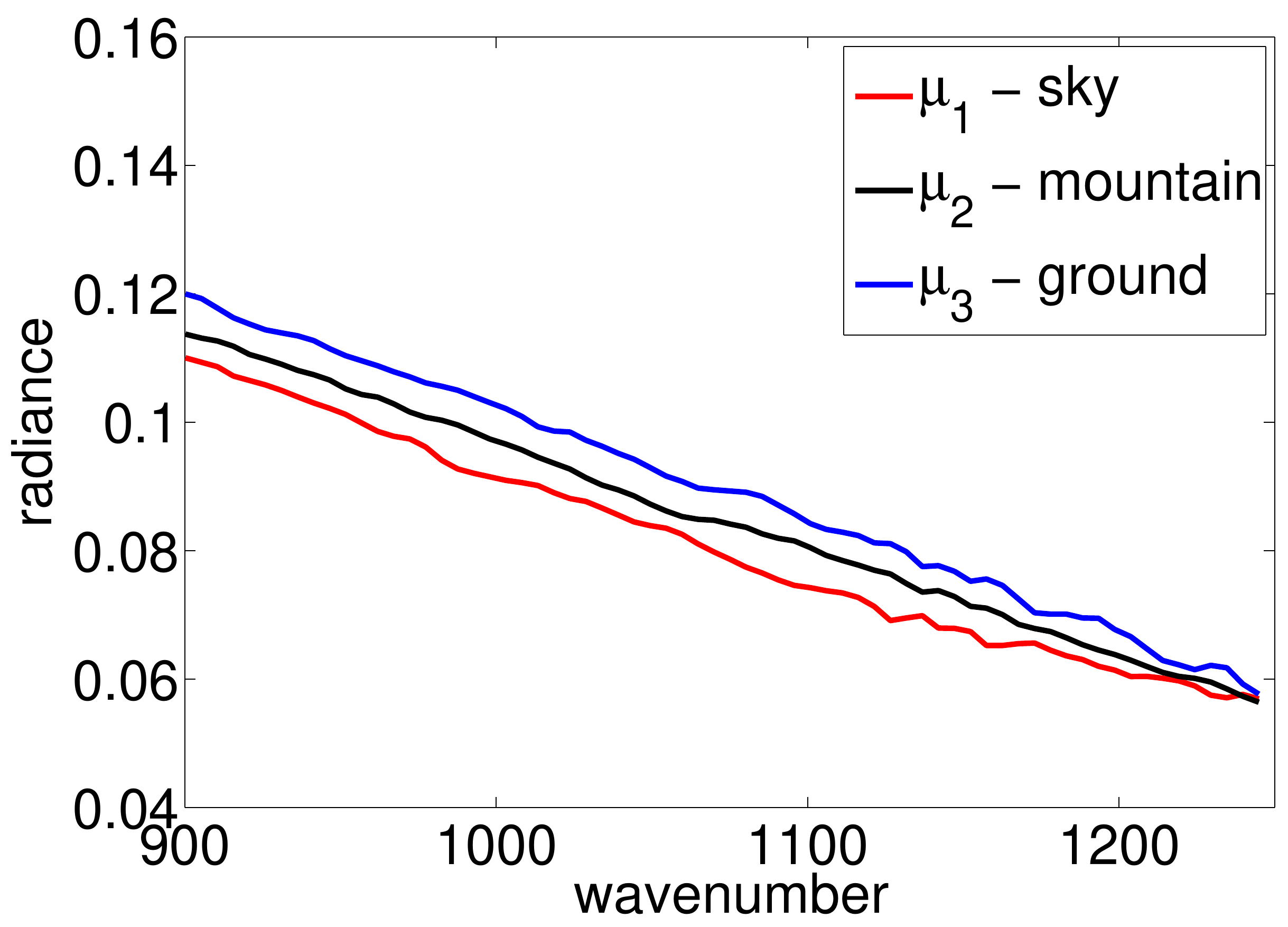}
\includegraphics[width=0.45\textwidth]{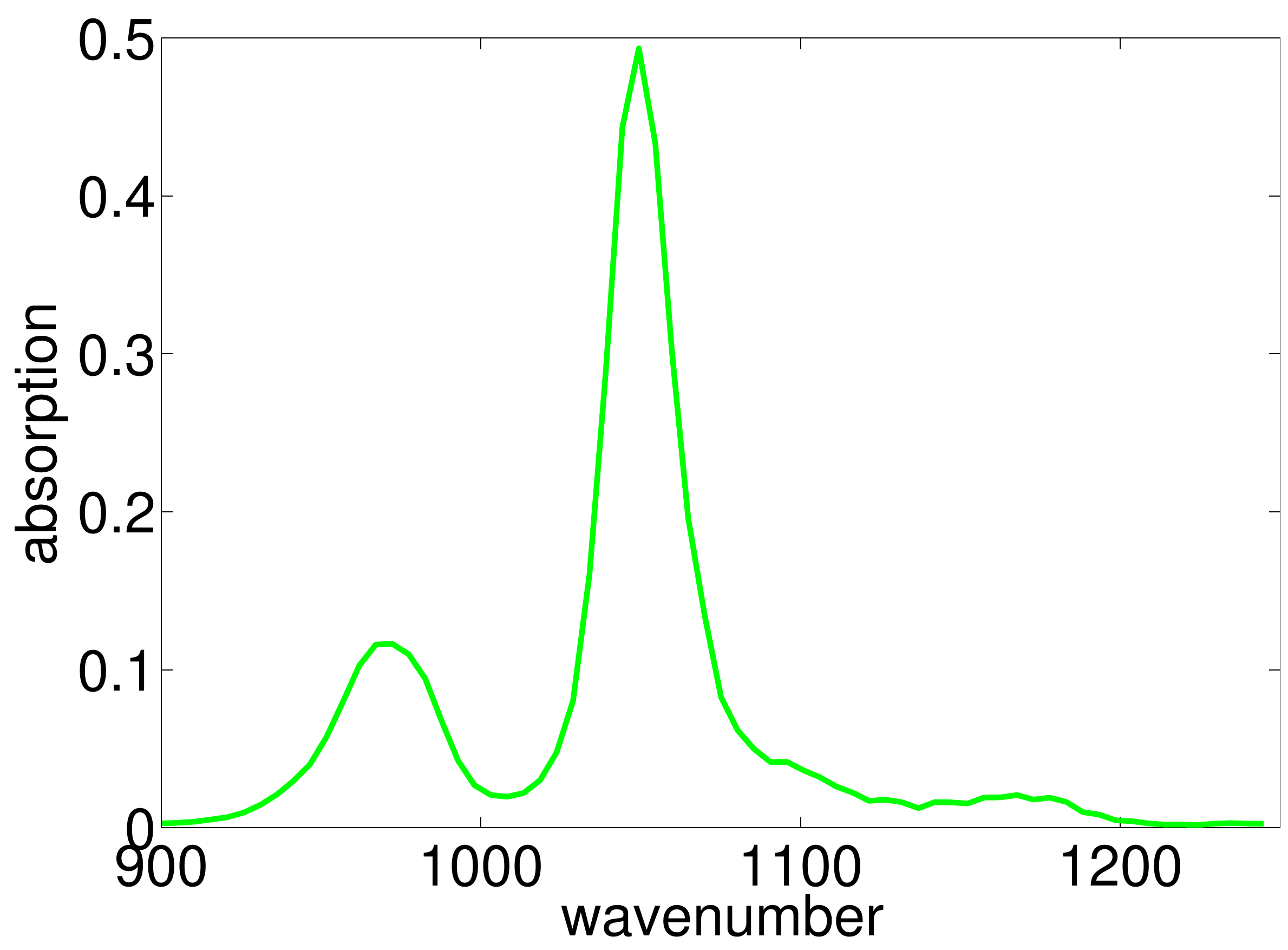}
\caption{Left: the mean spectra of sky, mountain and ground. Right: the absorption of the chemical plume. \label{fig:mus}}
\end{figure}

\noindent{\underline{\em{Description}}}.
We use synthetic data set to realize the Gaussian mixture model. We take the MIT Lincoln Lab Challenge Data (see details in the next section) to obtain the mean spectra for three regions - sky, mountain and ground (denoted by $\mu_1,\,\mu_2$ and $\mu_3$ respectively) and the target plume signature $\bs$. These spectra consist of 68 measurements at different values of wavelength and are shown in Figure~\ref{fig:mus}.

\begin{figure}[htbp!]
\centering
\includegraphics[width=0.45\textwidth]{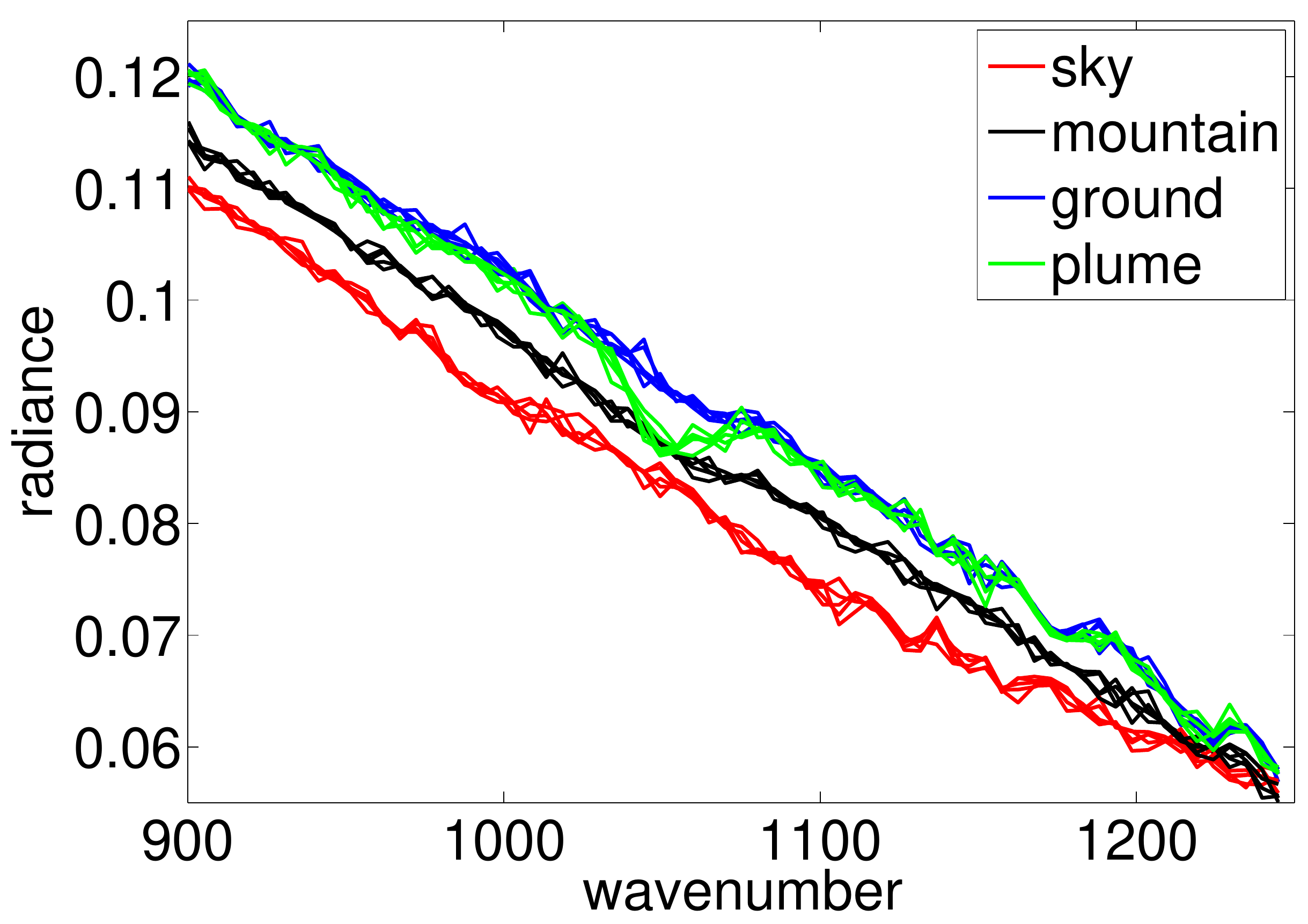}
\includegraphics[width=0.45\textwidth]{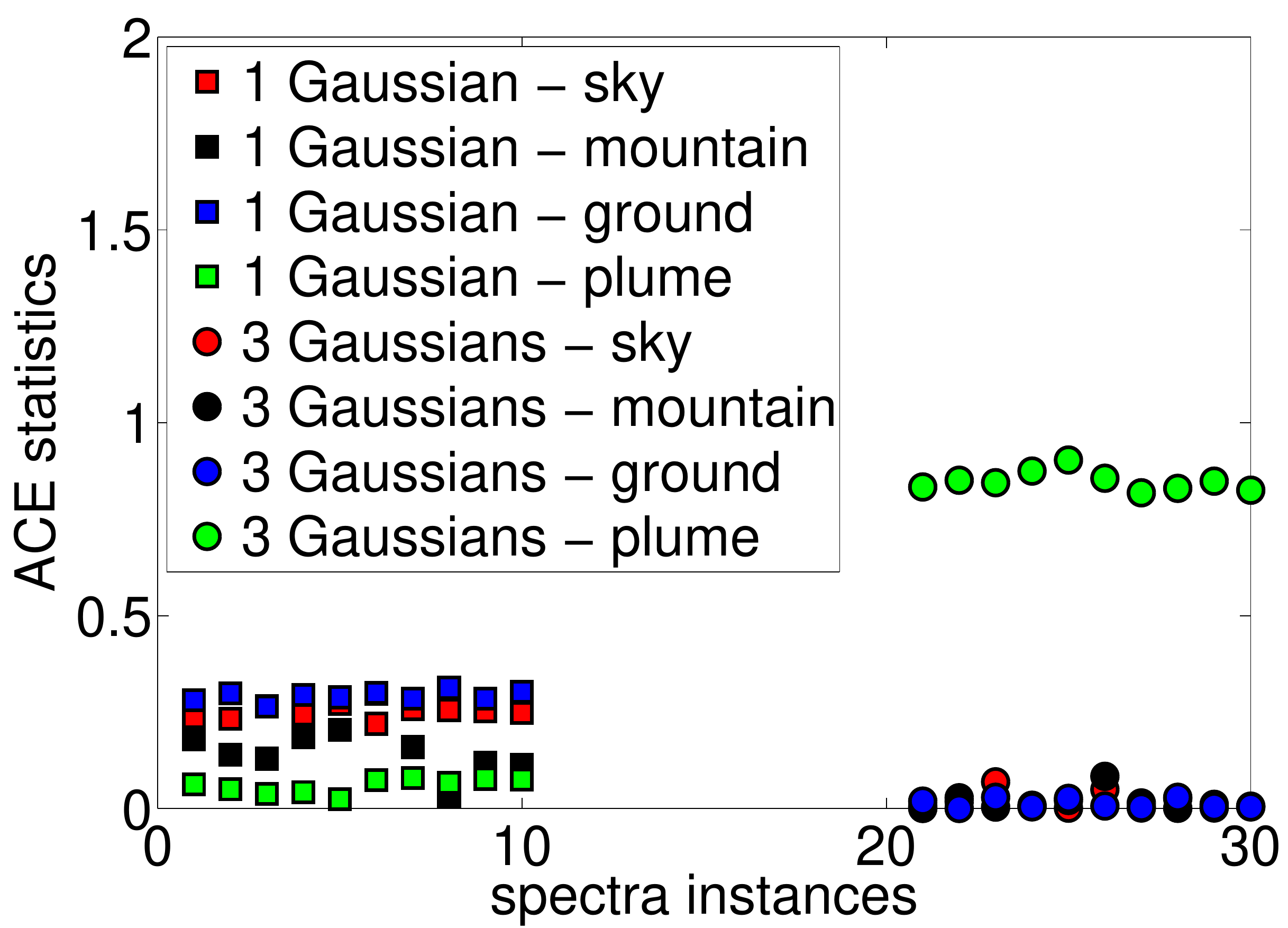}

\includegraphics[width=0.45\textwidth]{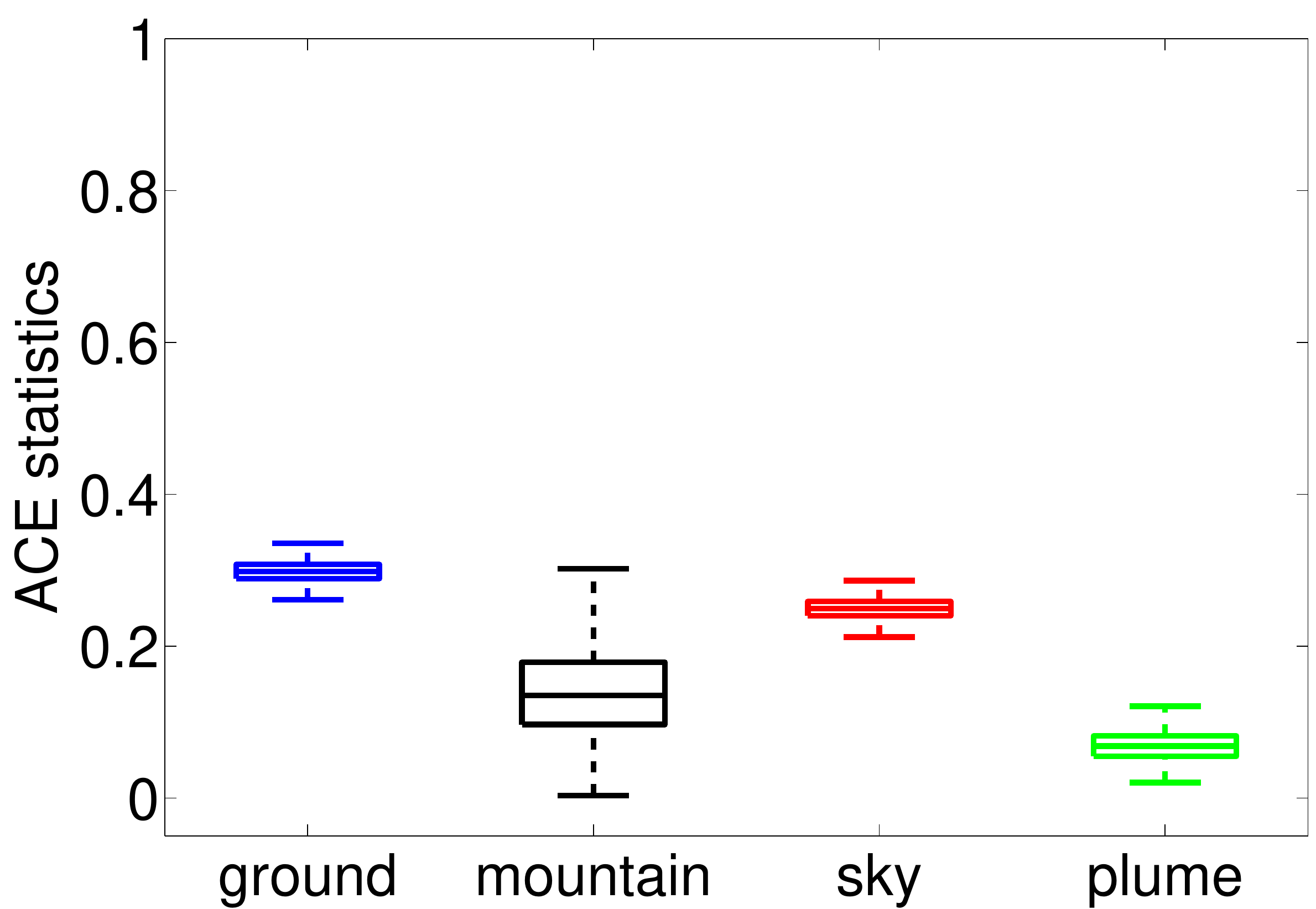}
\includegraphics[width=0.45\textwidth]{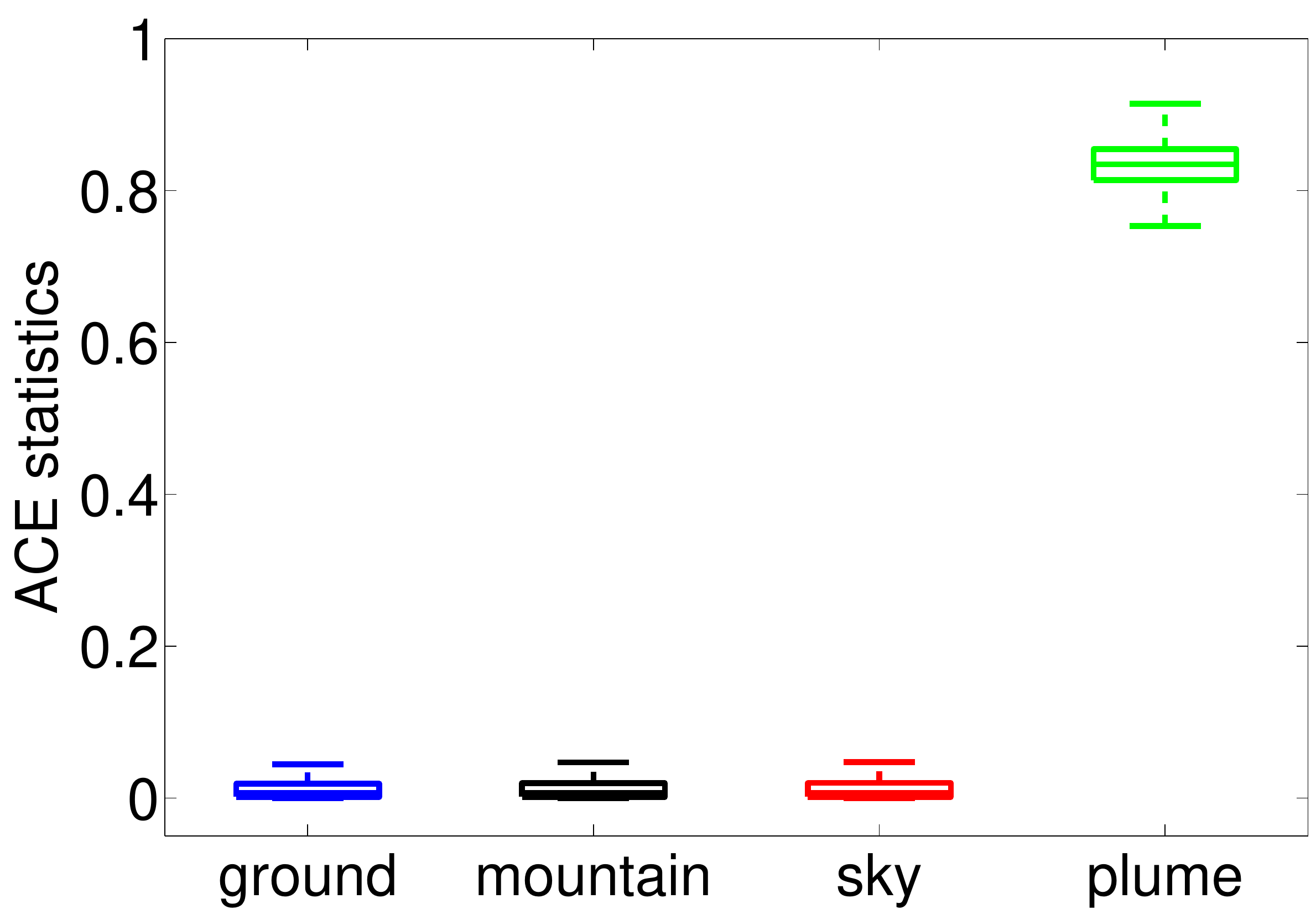}
\caption{Top left: radiance against wavenumber of 5 samples each from the groups of sky, mountain, ground and plume. Top right: the ACE scores of 10 spectra samples computed by 1 Gaussian and 3 Gaussians respectively. Bottom: the boxplot of the ACE scores of using 1 Gaussian (left) and 3 Gaussians (right). \label{fig:gaussian}}
\end{figure}

Moreover, $5,000,\,5,000$ and $4,000$ spectra are generated i.i.d. from Gaussian distributions $\cN(\mu_1,\Sigma_1)$, $\cN(\mu_2,\Sigma_2)$ and $\cN(\mu_3,\Sigma_3)$ respectively for the three regions, where $\Sigma_i = \diag\{\sigma_{i,1}^2,\cdots,\sigma_{i,68}^2\},\,i=1,2,3$ and all $\sigma_{i,j}$'s are drawn i.i.d. from the uniform distribution on $[0.002a,\, 0.008a]$ with $a=\max(\mu_3)$. Finally, $1,000$ spectra on the ground with chemical plume are generated from $g \bs+\bv$, where $\bv \sim \cN(\mu_3,\Sigma_3)$ and $g \sim \cN(-0.01,0.001)$. Top left of Figure~\ref{fig:gaussian} displays 5 sample spectra for each of the groups - sky, mountain, ground and plume.

\noindent{\underline{\em{Task}}. Detect the plume and compare the Gaussian mixture model with the single Gaussian model by the detection.

\noindent{\underline{\em{Technique}}}. We compute the ACE (or NMF) statistics (detection scores) using the ground truth. More precisely, we compute $T_{\text{mixNMF}}$ by~\eqref{eq:dect} using the true values of $\mu_i$ and $\Sigma_i$ with $i=1,\,2,\,3$ and $T_{\text{NMF}}$ by~\eqref{eq:NMF} using $\mu=(\mu_1+\mu_2+\mu_3)/3$ and $\Sigma = (\Sigma_1+\Sigma_2+\Sigma_3)/3$ for all spectra.

\noindent{\underline{\em{Results}}}.
Ten sample ACE statistics each from the 4 groups are demonstrated on the top right of Figure~\ref{fig:gaussian} for both the single Gaussian and mixture Gaussian models. In addition, the summary of these statistics are shown as boxplot at the bottom of Figure~\ref{fig:gaussian}, with the single Gaussian on the left and the mixture Gaussian on the right. The boxes are colored by groups of sky, mountain, ground and plume. On each box, the central mark is the median, the edges of the box are the 25th and 75th percentiles, the whiskers extend to the most extreme data points not considered as outliers. These figures show that when using three Gaussians instead of a single Gaussian, the ACE statistics of the plumes are greatly larger than those of the other regions. This makes the plumes more separable and thus more detectable.

\subsubsection{Subspace Mixture Models}

\noindent{\underline{\em{Description}}}.
We use synthetic data set to realize the subspace mixture model. $5,000,\,5,000$ and $4,000$ spectra are generated i.i.d. from $c\mu_1 + \epsilon$, $c\mu_2 + \epsilon$ and $c\mu_3 + \epsilon$ respectively for sky, mountain and ground regions, where $c \sim \cN(1,0.01)$, $\epsilon \sim \cN(0,\Sigma_{\epsilon})$, $\Sigma_{\epsilon} = \diag\{\sigma_{\epsilon,1}^2,\cdots,\sigma_{\epsilon,68}^2\}$ and $\sigma_{\epsilon,j}$'s are drawn i.i.d. from $\cN(0,0.005\max(\mu_3))$. Then $1,000$ spectra on the ground with plume are generated as $gs+c\mu_3 + \epsilon$ with $g \sim \cN(-0.01,0.001)$. Top left of Figure~\ref{fig:subspace} displays 5 sample spectra for each of the groups - sky, mountain, ground and plume.

\noindent{\underline{\em{Task}}. Detect the plume and compare the subspace mixture model with the single subspace model by the detection.

\noindent{\underline{\em{Technique}}}. We compute the NSS statistics (detection scores) using the ground truth. More precisely, we compute $T_{\text{mixNSS}}$ by~\eqref{eq:dect} letting $B_i = \mu_i$, $i=1,\,2,\,3$ and $T_{\text{NSS}}$ by~\eqref{eq:NSS} using $B = (\mu_1 + \mu_2 + \mu_3) / 3 $.

\noindent{\underline{\em{Results}}}.
Ten sample NSS statistics each from the 4 groups are demonstrated on the top right of Figure~\ref{fig:subspace} for both the single subspace and mixture subspace models. In addition, the summary of these statistics are shown as boxplot at the bottom of Figure~\ref{fig:subspace}, with the single subspace on the left and the mixture subspace on the right. From these figures, we know that the separation between the plume and other regions greatly improves when using the mixture subspace model.

\begin{figure}[htbp!]
\centering
\includegraphics[width=0.45\textwidth]{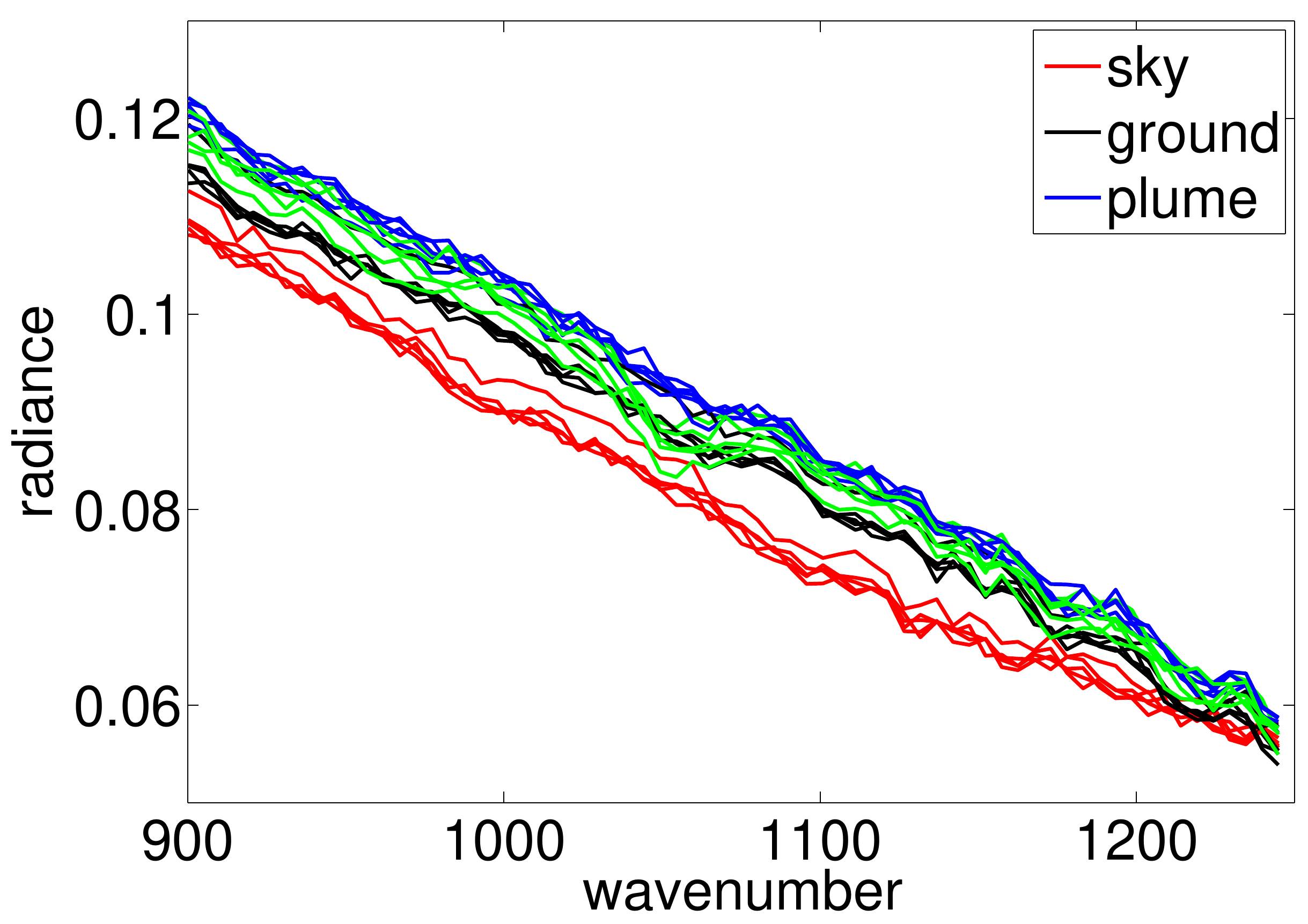}
\includegraphics[width=0.45\textwidth]{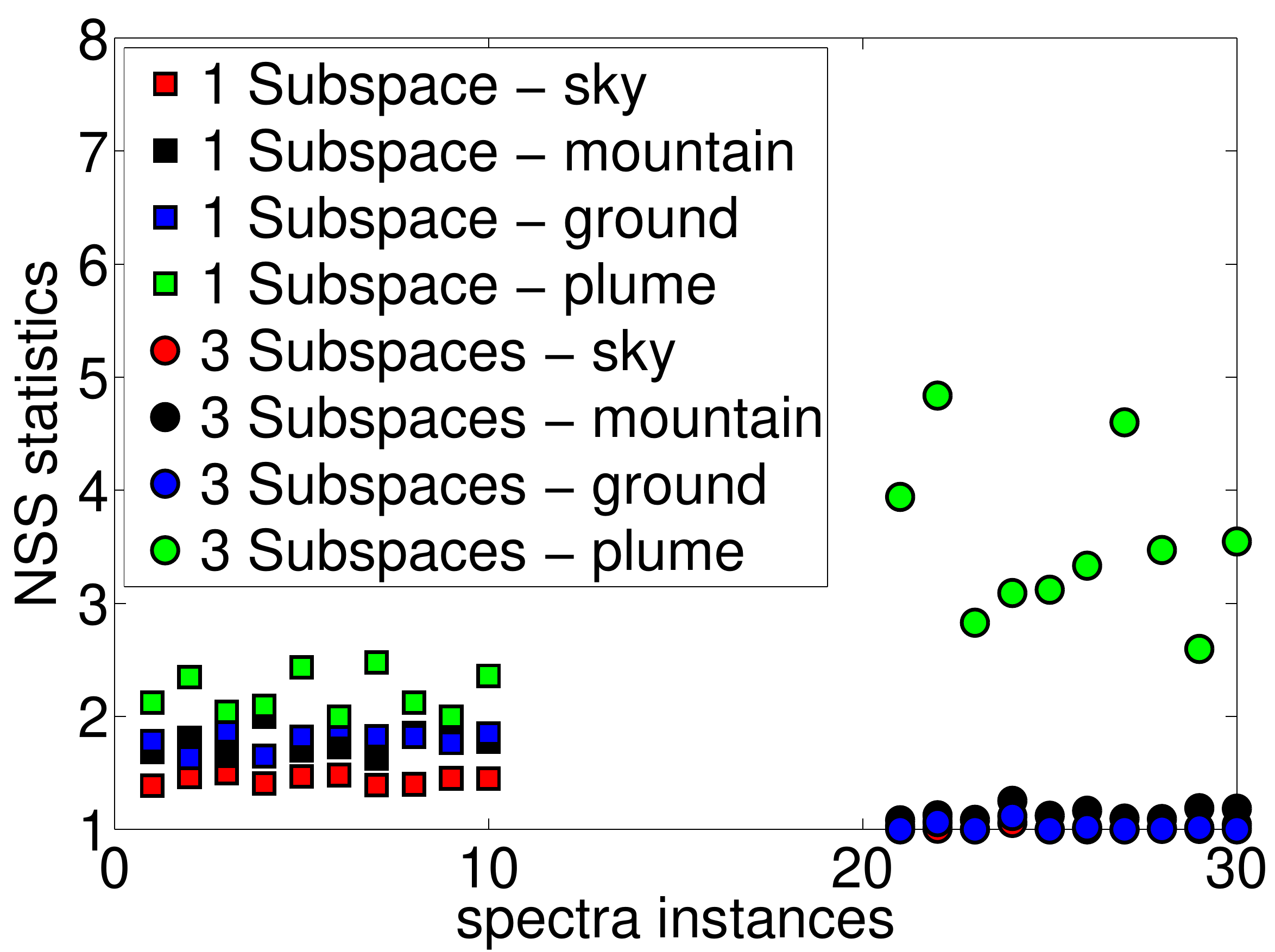}

\includegraphics[width=0.45\textwidth]{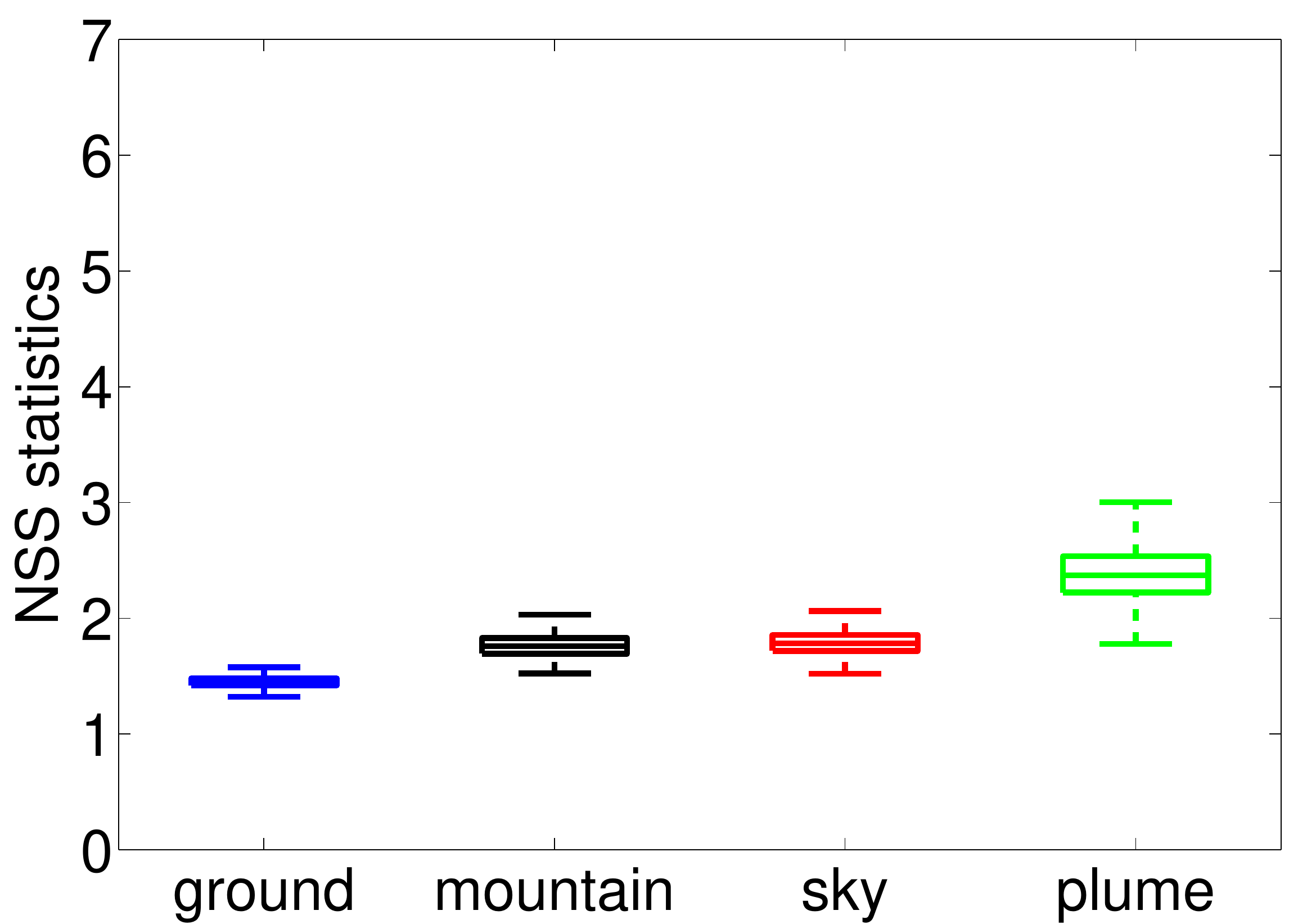}
\includegraphics[width=0.45\textwidth]{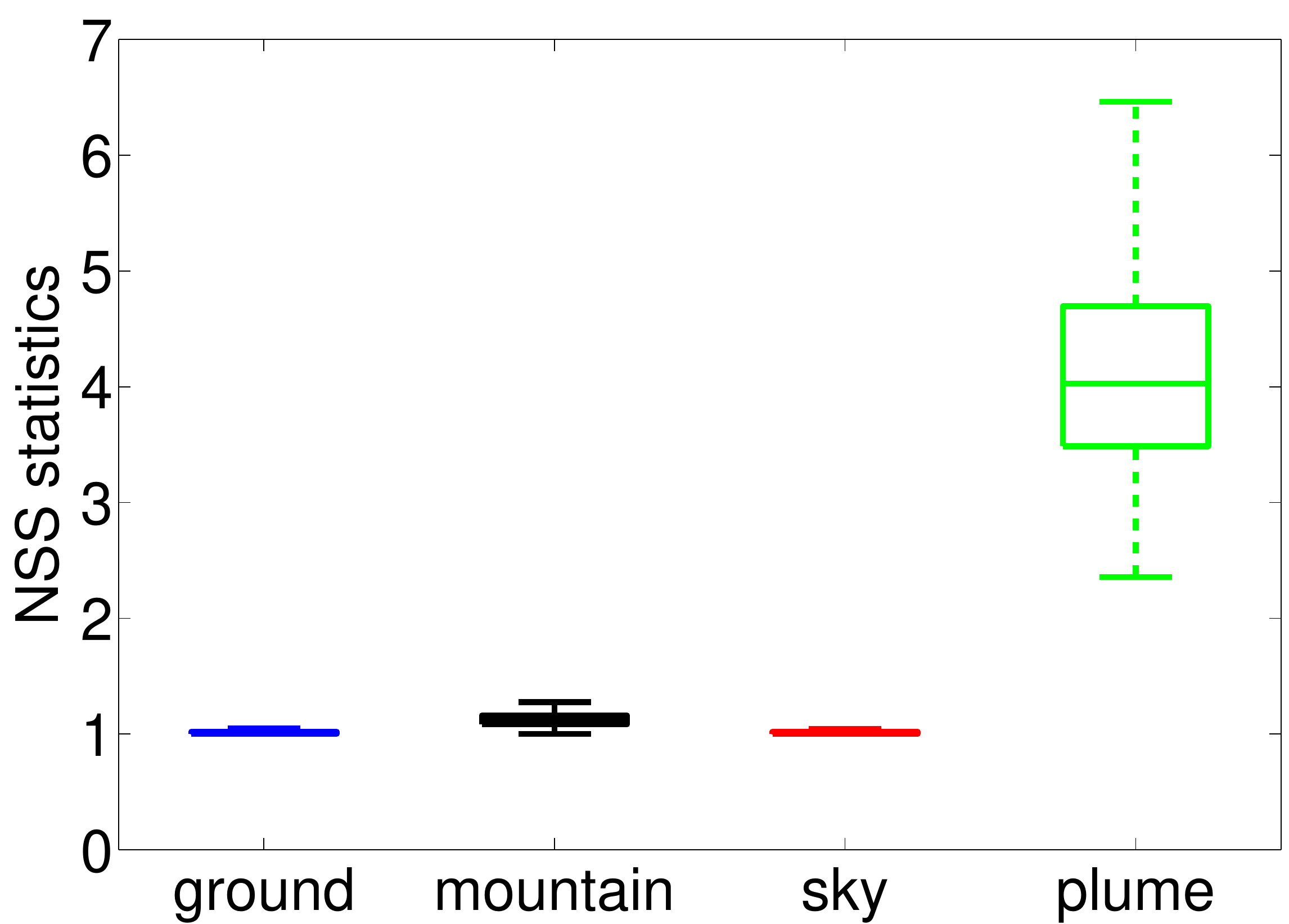}
\caption{Top left: radiance against wavenumber of 5 samples each from the groups of sky, mountain, ground and plume. Top right: the NSS scores of 10 samples computed by 1 subspace and 3 subspaces respectively. Bottom: the boxplot of the NSS scores of using 1 Gaussian (left) and 3 Gaussians (right). \label{fig:subspace}}
\end{figure}

\subsubsection{Partial Least Square Regression}
\noindent{\underline{\em{Description}}}. We generate $10,000$ spectra from $\mu+b \bfe$, where $\mu=(\mu_1+\mu_2+\mu_3)/3$ , $b=\max{\mu}$ and $\bfe_j \sim \text{Poisson}(0.005)$, i.i.d., $j=1,\cdots,68$. Then $1,000$ spectra are generated from $\mu+b\bfe+g\bs$ with $g \sim \cN(-0.01,0.001)$. 10 sample spectra of each class (scene and plume) are shown in Figure~\ref{fig:plsr1}.
\begin{figure}[htbp!]
\centering
\includegraphics[width=0.8\textwidth]{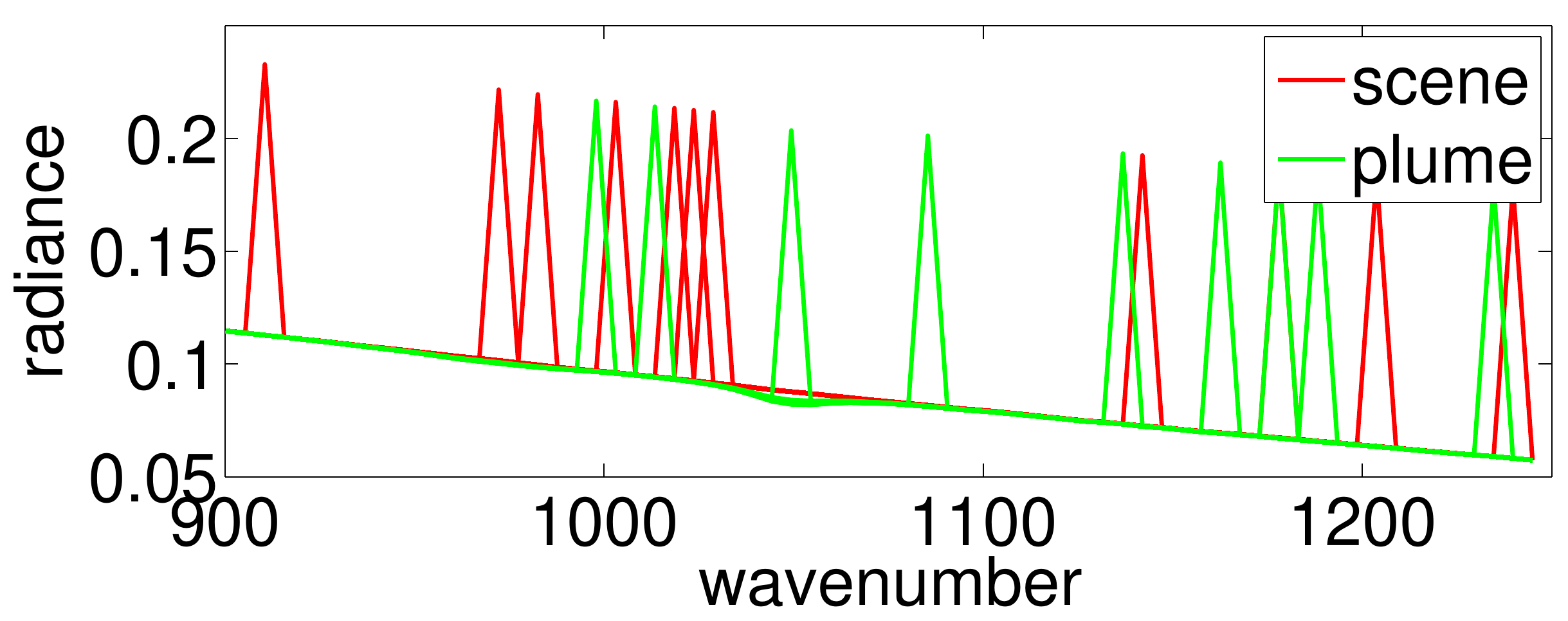}
\caption{Background and plume spectra contaminated by Poisson noise.}
\label{fig:plsr1}

\end{figure}

\begin{figure}[htbp!]
\centering
\includegraphics[width=0.45\textwidth]{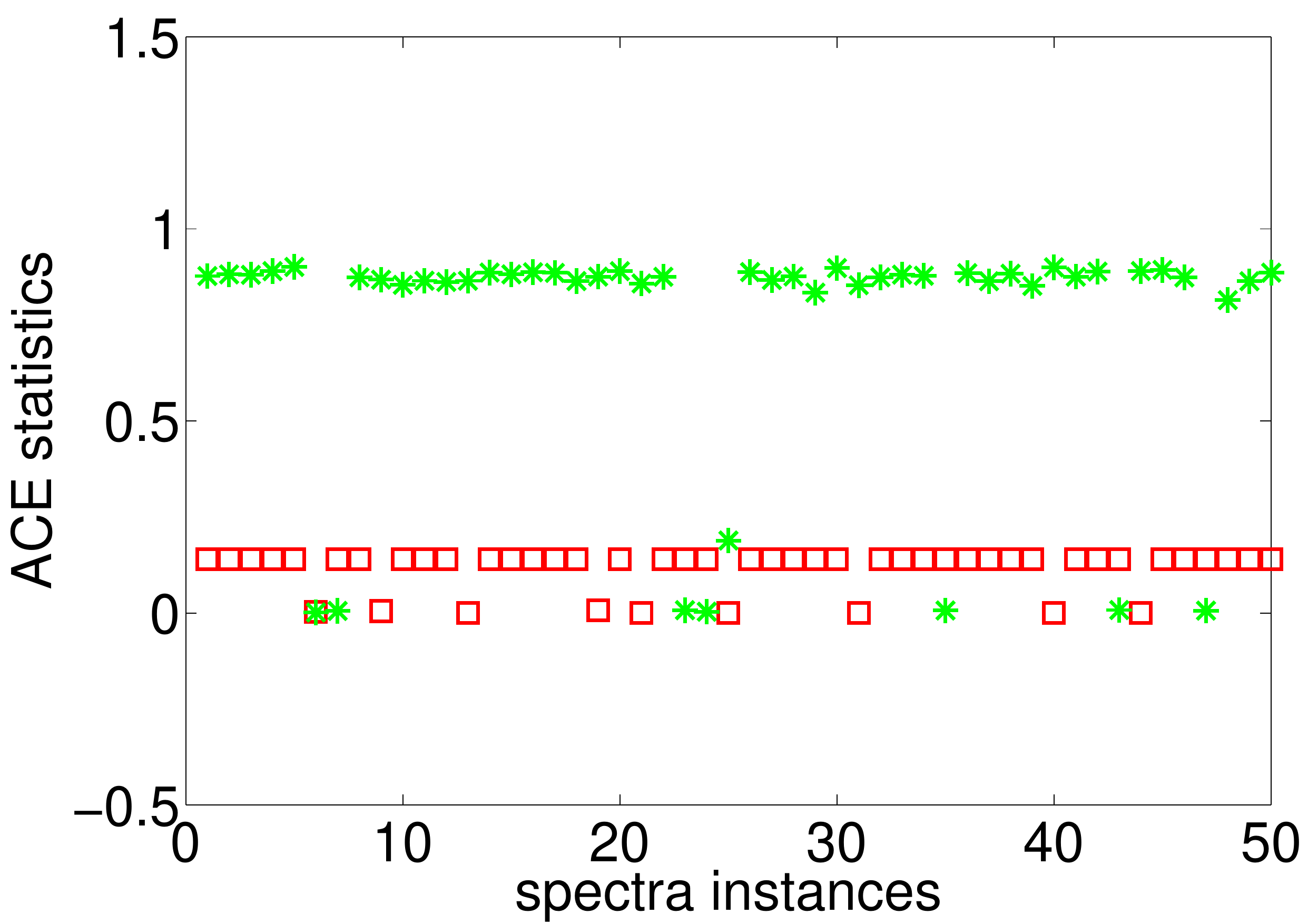}
\includegraphics[width=0.45\textwidth]{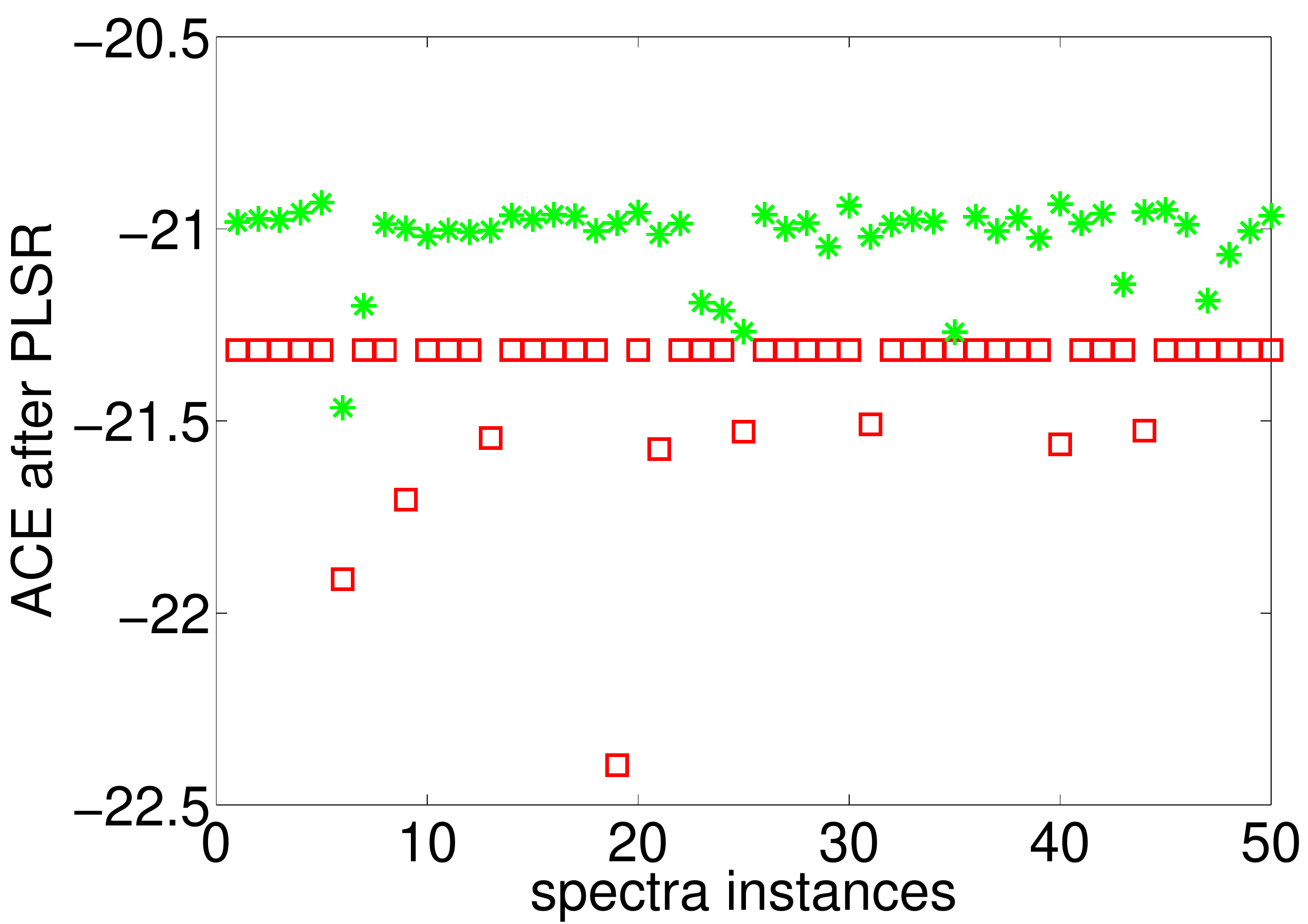}

\includegraphics[width=0.45\textwidth]{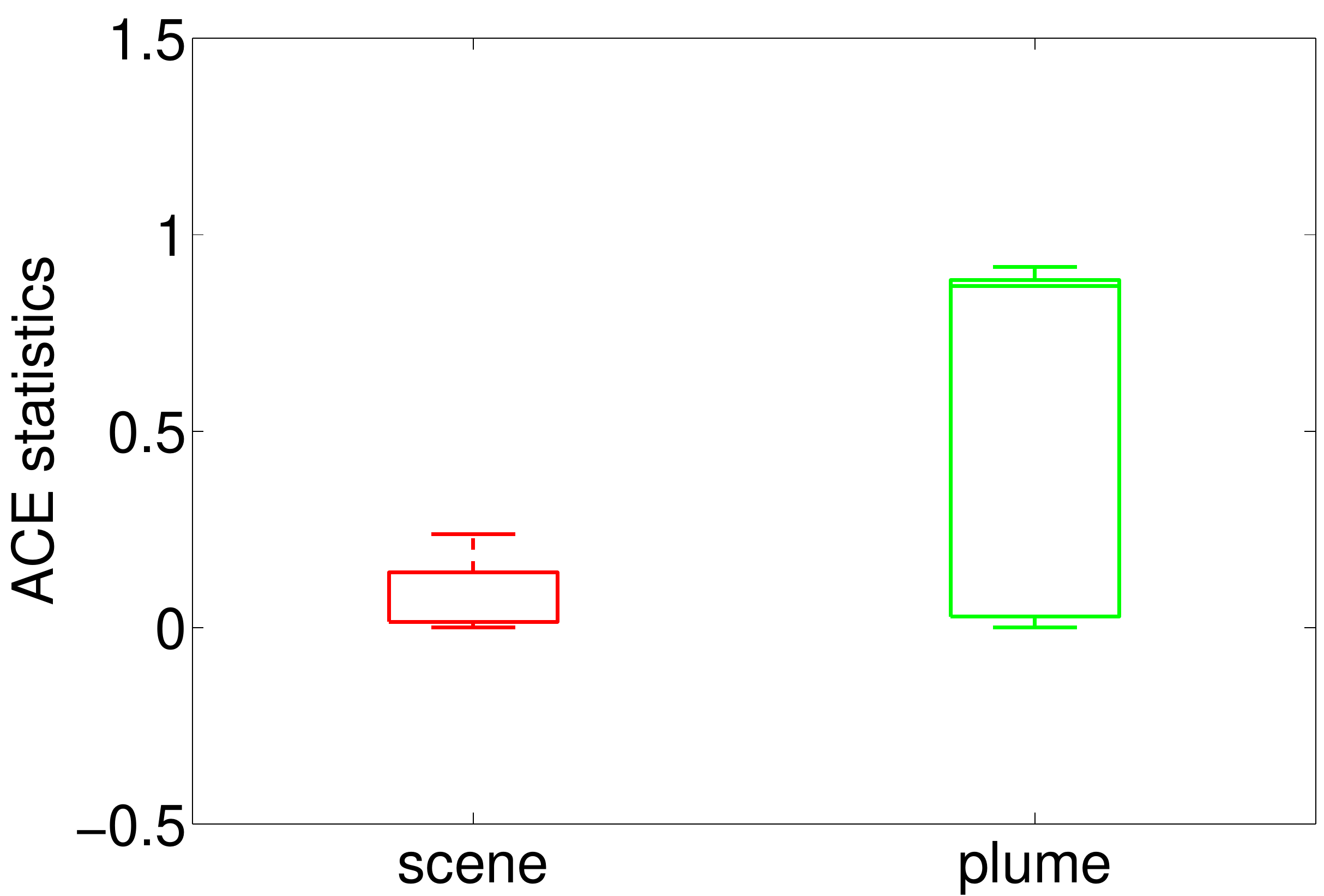}
\includegraphics[width=0.45\textwidth]{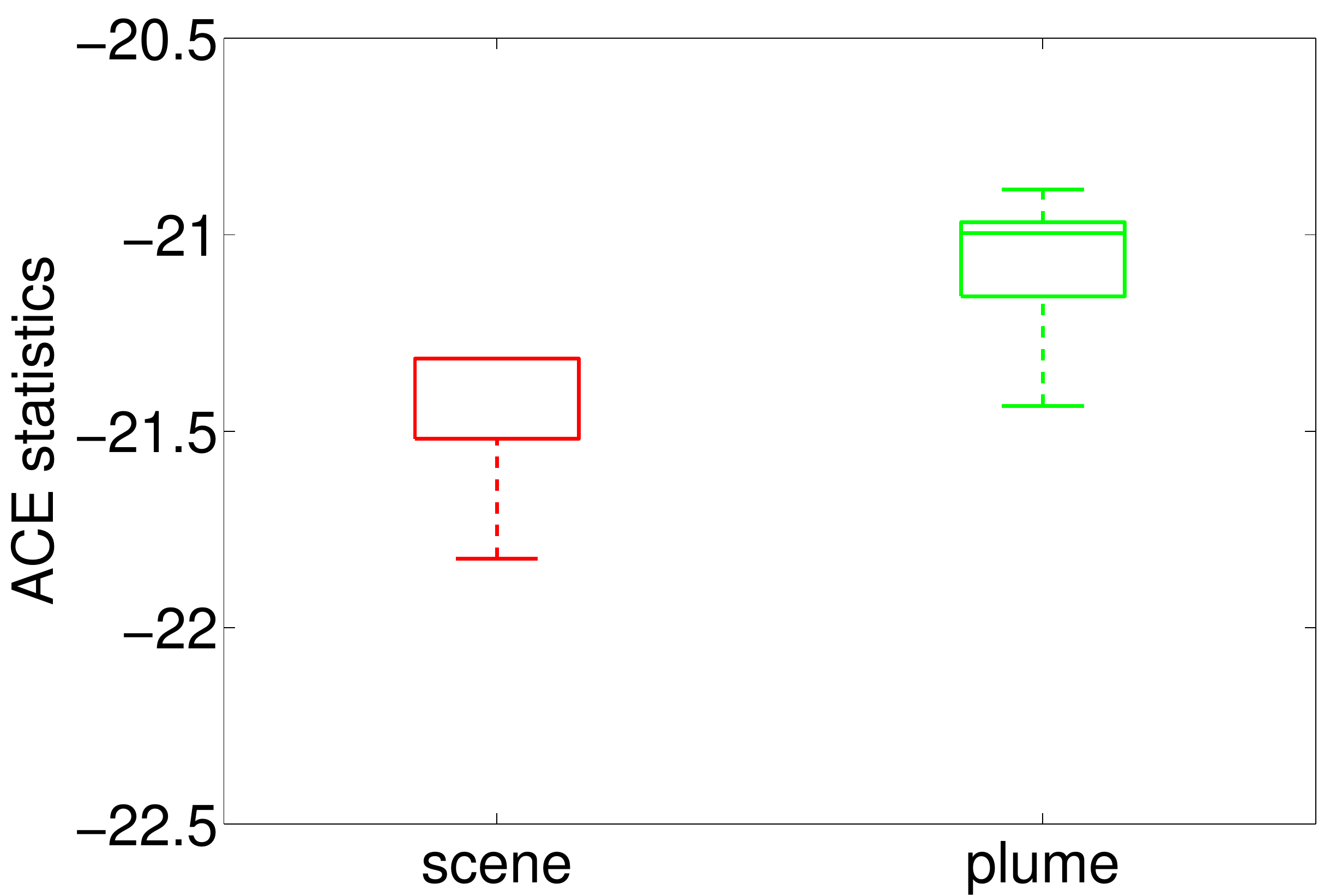}
\caption{Top left: ACE statistics of 50 sample spectra of each class; Top right: ACE statistics after PLSR of the same 50 sample spectra of each class; Bottom left: the boxplot of the ACE statistics; Bottom right: the boxplot of the ACE statistics with PLSR. \label{fig:plsr}}

\end{figure}

\noindent{\underline{\em{Task}}. Detect the plume and compare the performances before and after PLSR.

\noindent{\underline{\em{Technique}}}. We compute the ACE statistics as described in Section~\ref{sec:prev}. Then we apply Algorithm~\ref{alg:plsr} with $\tau_2=\tau_3=0.2$ to the computed ACE scores.

\noindent{\underline{\em{Results}}}.
The top of Figure~\ref{fig:plsr} displays the 50 sample ACE scores on the left and the corresponding enhanced results after applying PLSR on the right. Their statistical summary are demonstrated as boxplots at the bottom of Figure~\ref{fig:plsr} with ACE on the left and PLSR on the right. These figures demonstrate that the PLSR procedure greatly improves the separation of the detection scores between scene and plume.

\subsubsection{Multiple Chemical Plumes}
\noindent{\underline{\em{Description}}}. We follow Section~\ref{sec:GaussData} to generate $5,000$ sample spectra each for the sky, mountain and ground areas, then we generate $1,000$ sample spectra for the ground area each with chemical plume 1 and 2. The signature ($\bs_1,\bs_2$) of these two plumes are demonstrated on the top left of Figure~\ref{fig:2plumes}. Finally we generate 100 sample spectra for the ground area with both chemical plumes from $g_1 \bs_1+g_2\bs_2+\bv$, where $\bv \sim \cN(\mu_3,\Sigma_3)$ and $g_1,g_2 \sim \cN(-0.01,0.001)$. On the top right of Figure~\ref{fig:2plumes}, sample spectra of different areas are displayed. 

\begin{figure}[htb!]
\centering
\includegraphics[width=0.4\textwidth]{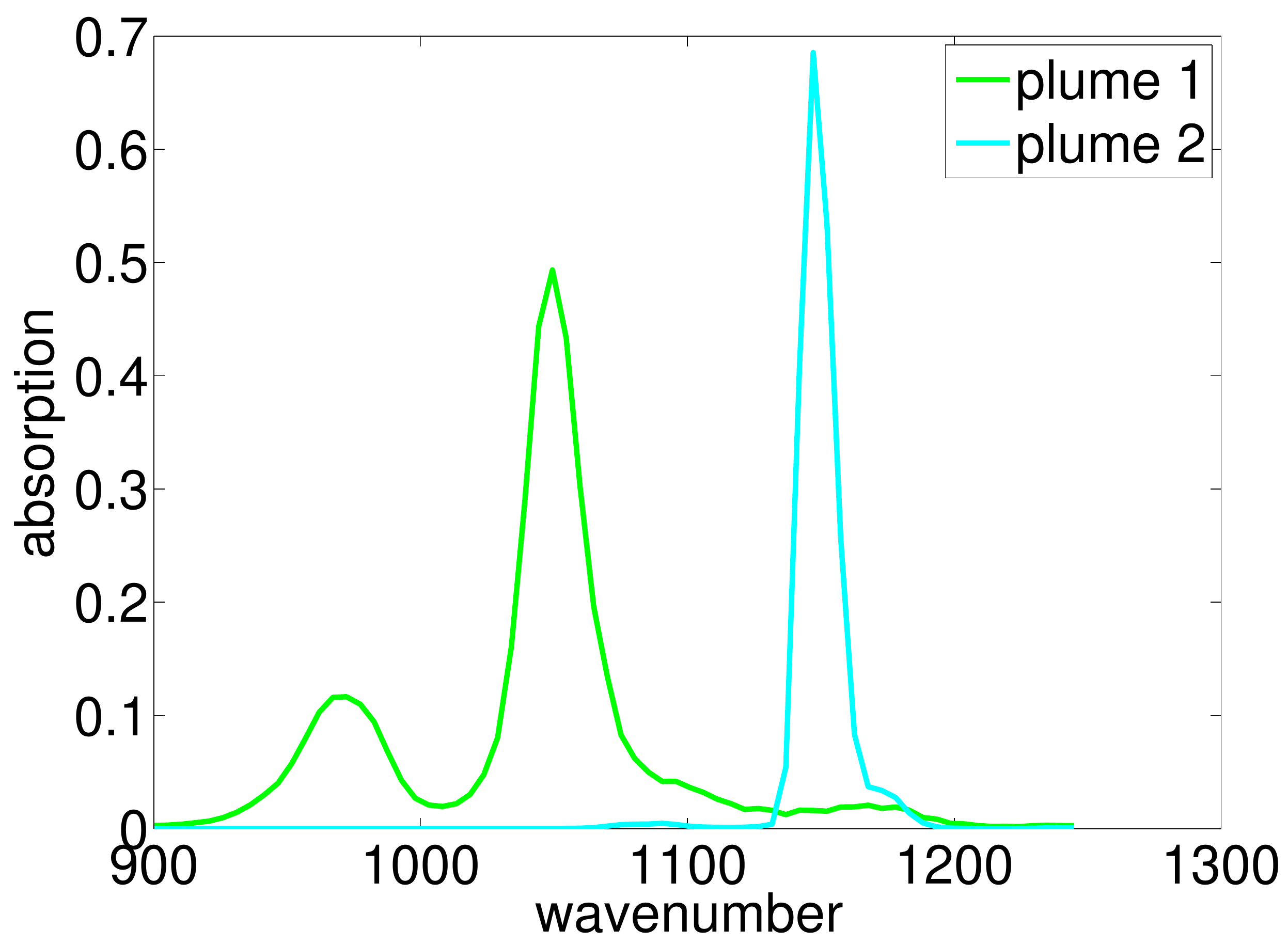}
\includegraphics[width=0.4\textwidth]{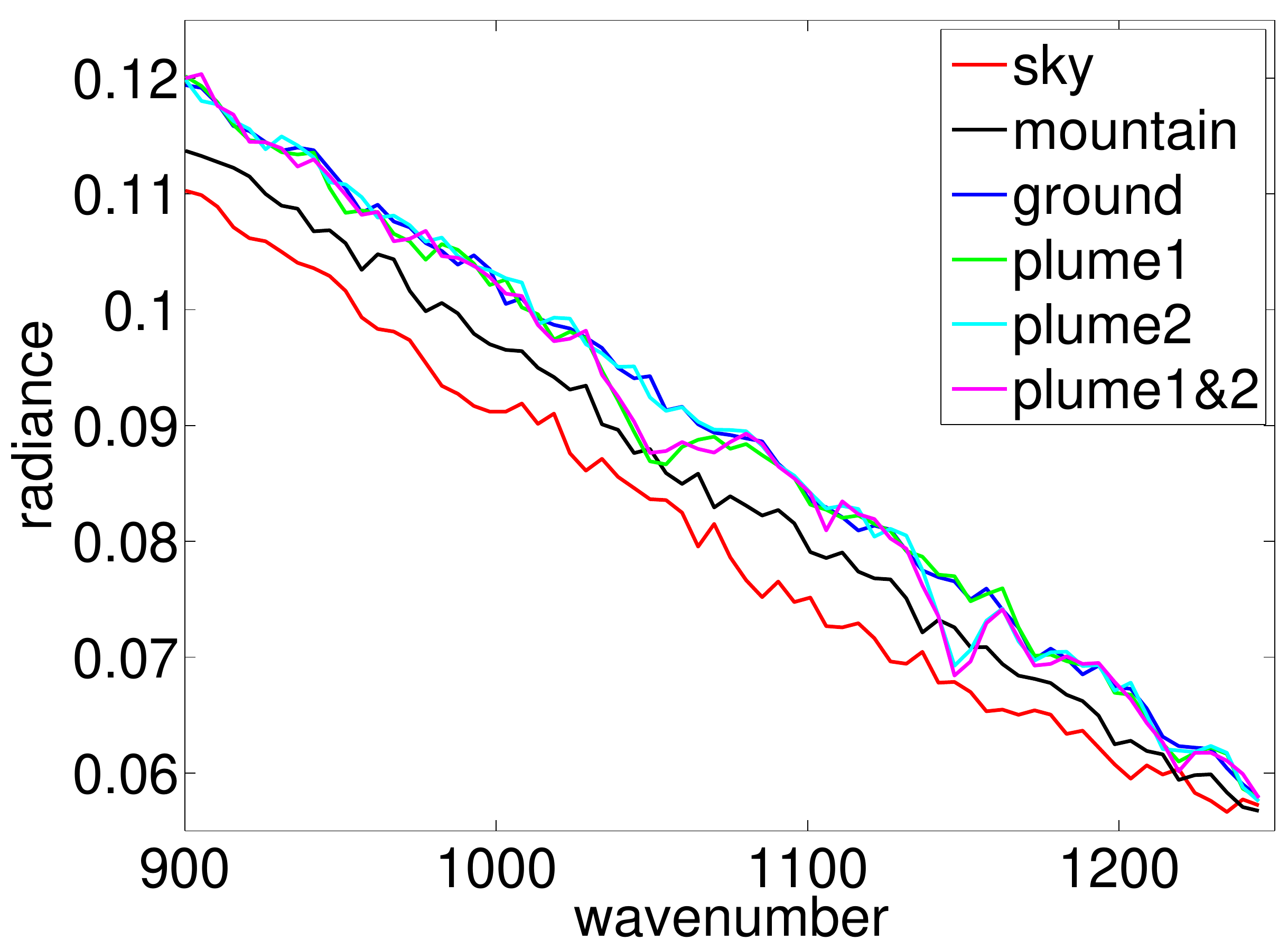}

\includegraphics[width=0.4\textwidth]{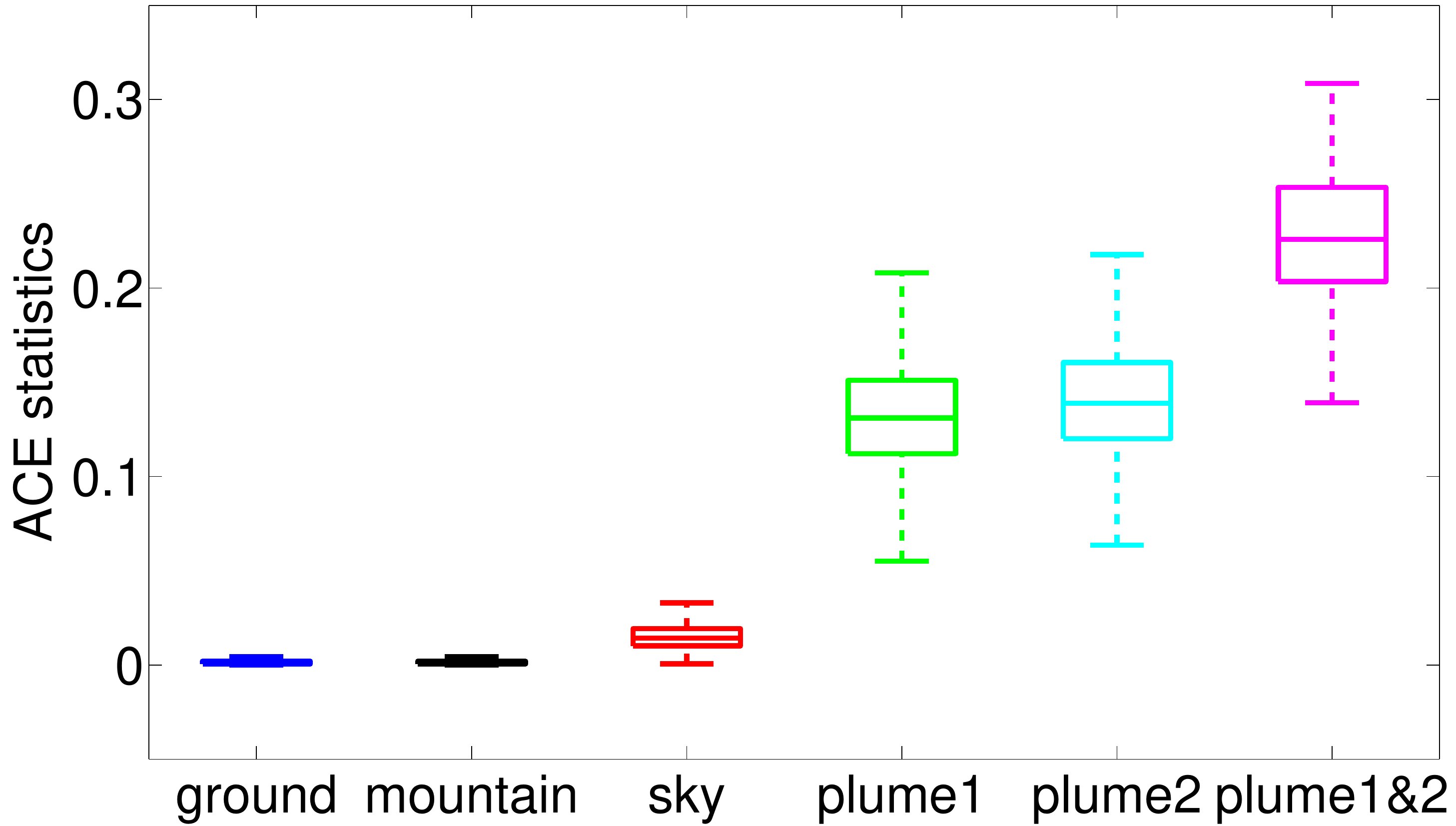}
\includegraphics[width=0.4\textwidth]{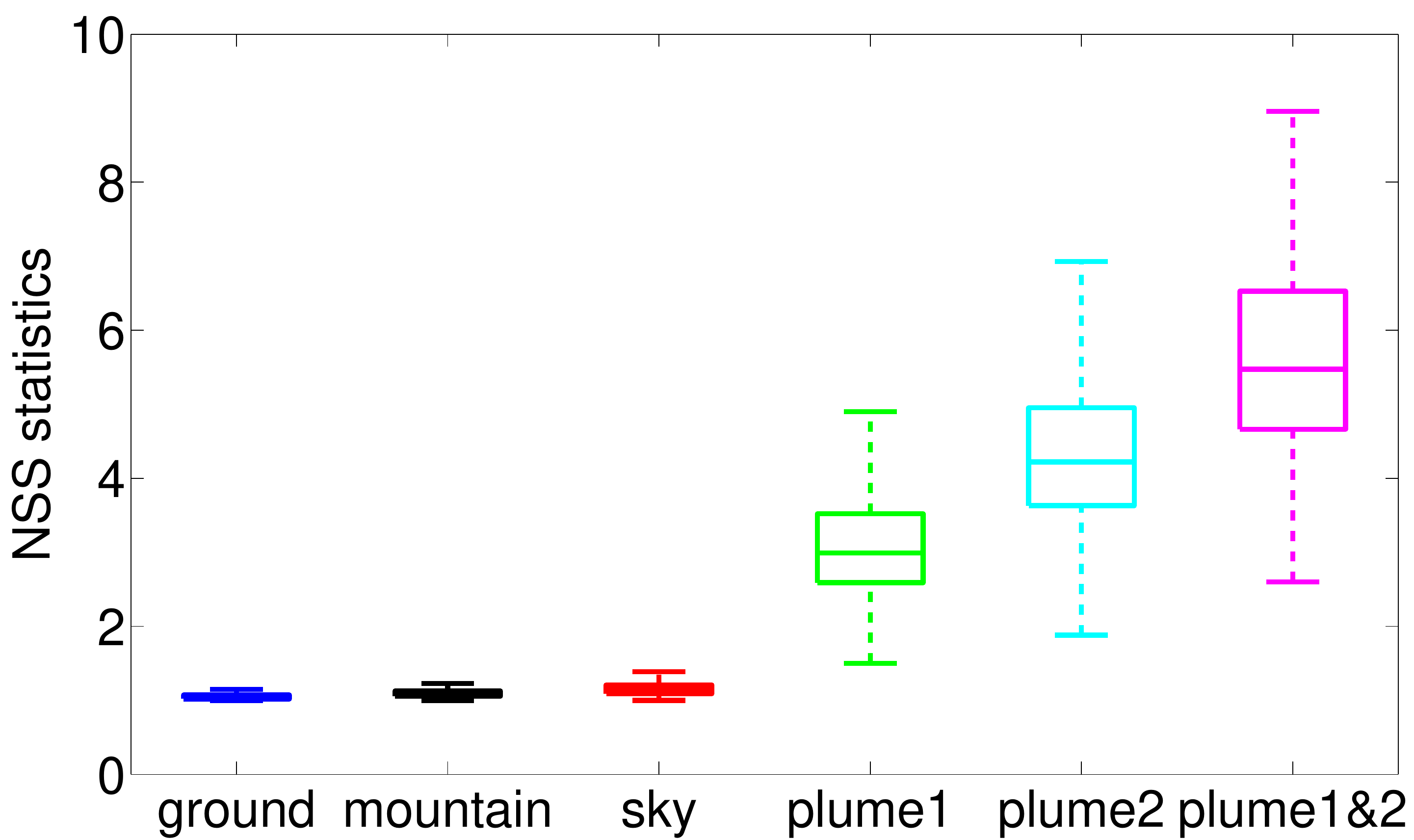}
\caption{Top left: signature spectra of plume 1 and plume 2. Top right: sample spectra of different areas, with and without plumes. Bottom left: boxplot of the ACE statistics for different areas. Bottom right: boxplot of the NSS statistics for different areas.\label{fig:2plumes}}
\end{figure}

\noindent{\underline{\em{Task}}. Detect where a plume (or both plumes) are present.

\noindent{\underline{\em{Technique}}}. We compute the detectors as described in Section~\ref{sec:multiPlume}.

\noindent{\underline{\em{Results}}}. The box plots of the NMF (ACE) statistics and the NSS statistics are shown at the bottom of Figure~\ref{fig:2plumes}. It is evident that the regions with plumes can be distinguished from those without plumes very well. We will not conduct an experimental analysis as careful as for a single target plume for two reasons. First, all the real data under consideration have only a single chemical plume. Second, from the statistical point of view, our models (both Gaussian and subspace) for different numbers of chemical plumes are essentially the same models with different dimensions. We will just use this simple example in this section to justify that our methods work effectively for $N>1$.

\subsection{MIT Lincoln Lab Challenge Data\label{sec:MITdata}}
\noindent{\underline{\em{Description}}}. We use hyperspectral images available at the CSR, collected by the MIT Lincoln Lab, and made available through the ATD program. This data set consists of four individual hyperspectral images, two of which are for released chemical plume and two for embedded plume. In each of the four, available data include a radiance data cube, its matrix form, the absorption coefficient spectrum of the chemical plume of interest and plume present mask. Data cubes are about of the size $200\times300\times100$, where $200\times300$ is the spatial size and $100$ is the spectral dimension.

\noindent{\underline{\em{Task}}. Detect the chemical plume from a single cube.

\noindent{\underline{\em{Technique}}}. We apply the three detection algorithms (NMF, NSS, LC), their corresponding mixture models (mixNMF, mixNSS, mixLC), and the two enhancement techniques (resampling and PLSR) as described in Section~\ref{sec:prev} and~\ref{sec:alg}. The dimension for the subspaces is chosen to be $2$ by looking at where the singular values start to flatten. The parameters for the enhancement techniques are $\tau_1=0.2,\tau_2 =\tau_3=0.15$.

\noindent{\underline{\em{Results}}}. The receiver operating characteristic (ROC) curves are shown in Figure~\ref{fig:MIT_eg4} for one of the embedded data cube (also the most difficult one). Comparisons are made between single and mixture models on the left and between mixture models and those followed by resampling and then PLSR (the combination of these two enhancement techniques are denoted as {\it -eh}) on the right. From the left figure, we see that using mixture models for background can improve the detection results. Among the three algorithms, mixture LC works the best. In the right figure, when adding the enhancement procedures, both mixNMF and mixNSS improve, but mixLC does not. Overall, mixNMF with enhancement outperforms others.

\begin{figure}[htbp!]

\centering
\includegraphics[width=0.45\textwidth]{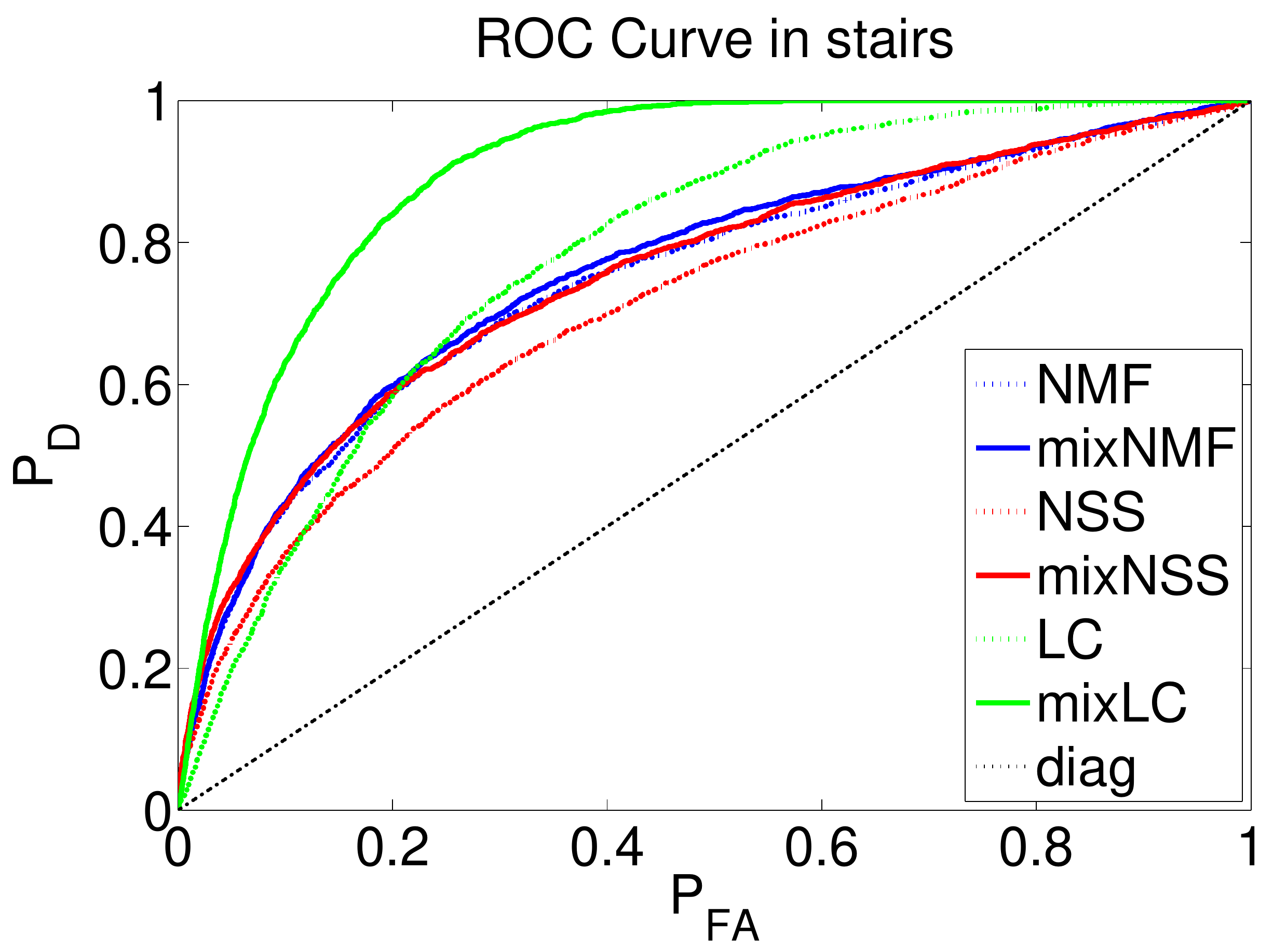}
\includegraphics[width=0.45\textwidth]{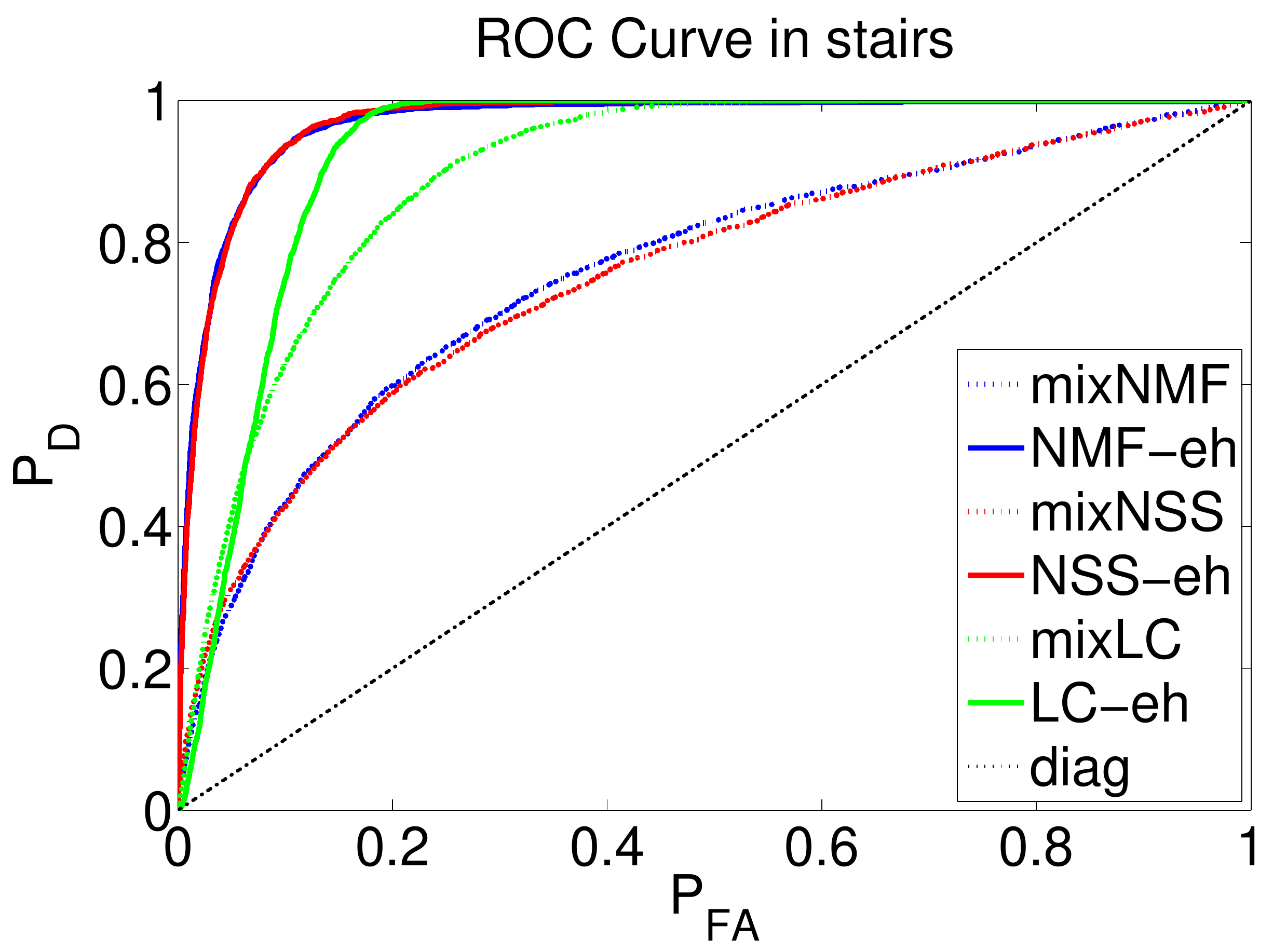}
\caption{ROC curves. The left is comparison for single and mixture models and the right for mixture models before and after enhancement (denoted as {\it -eh}). \label{fig:MIT_eg4}}

\end{figure}

Furthermore, we demonstrate the detection maps of the intermediate steps of the best performed method mixNMF{\it -eh}, namely mixNMF, mixNMF with resampling (mixNMF-rs), mixNMF with resampling twice (mixNMF-rs2), mixNMF with resampling twice and followed by PLSR (mixNMF-rs2-plsr). These results are shown with those of the NMF on the original data (NMF-outliers) and on the data with outliers removed (NMF). All the methods stated in this section proceed the outlier removal first, except NMF-outliers. Figure \ref{fig:MIT_eg4_steps} shows that the mixture model and the various enhancement steps (resampling and PLSR) improve the results gradually; the plume region becomes more and more separable over these steps.

\begin{figure}[htbp!]

\centering
\includegraphics[width=0.45\textwidth]{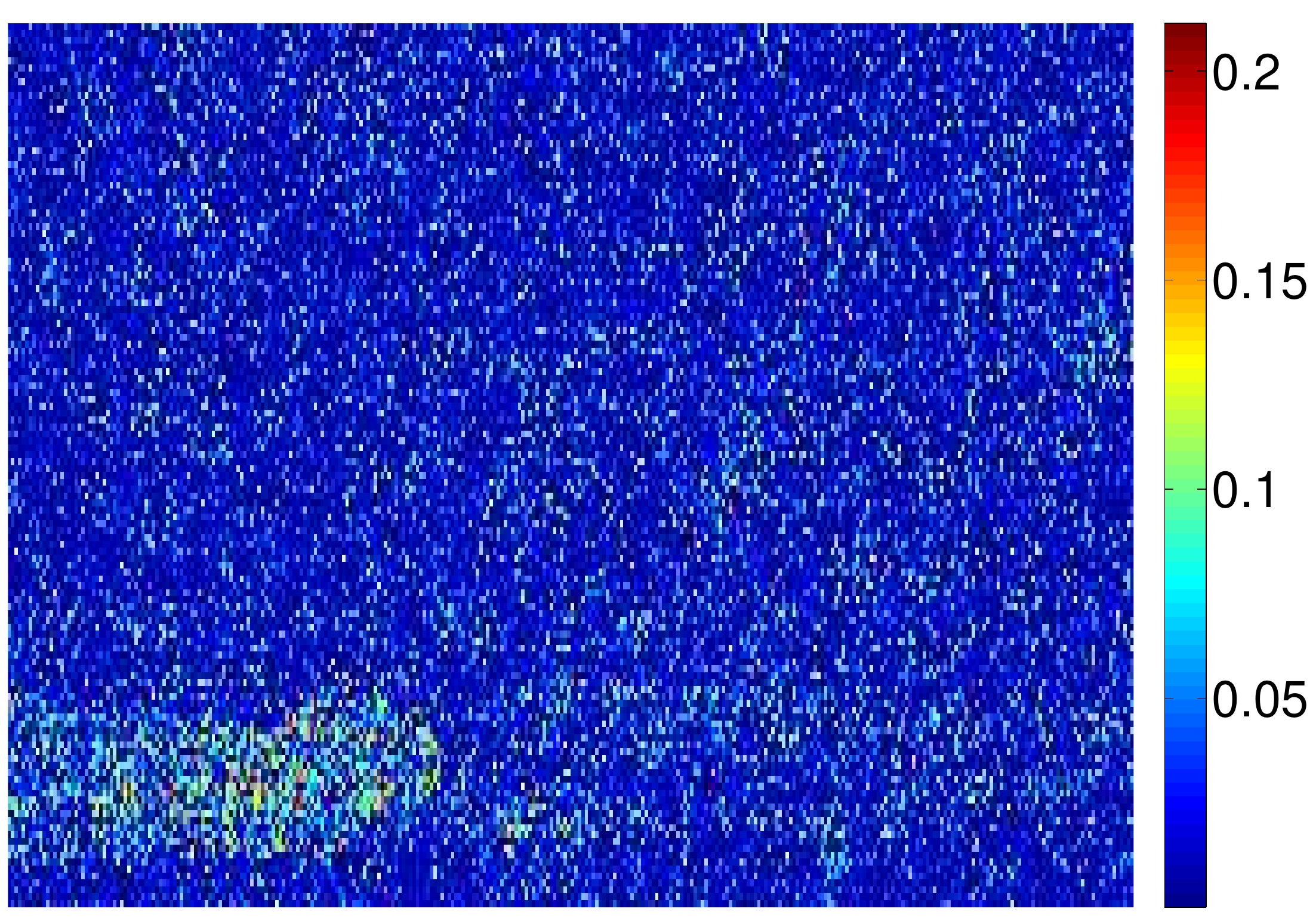}
\includegraphics[width=0.45\textwidth]{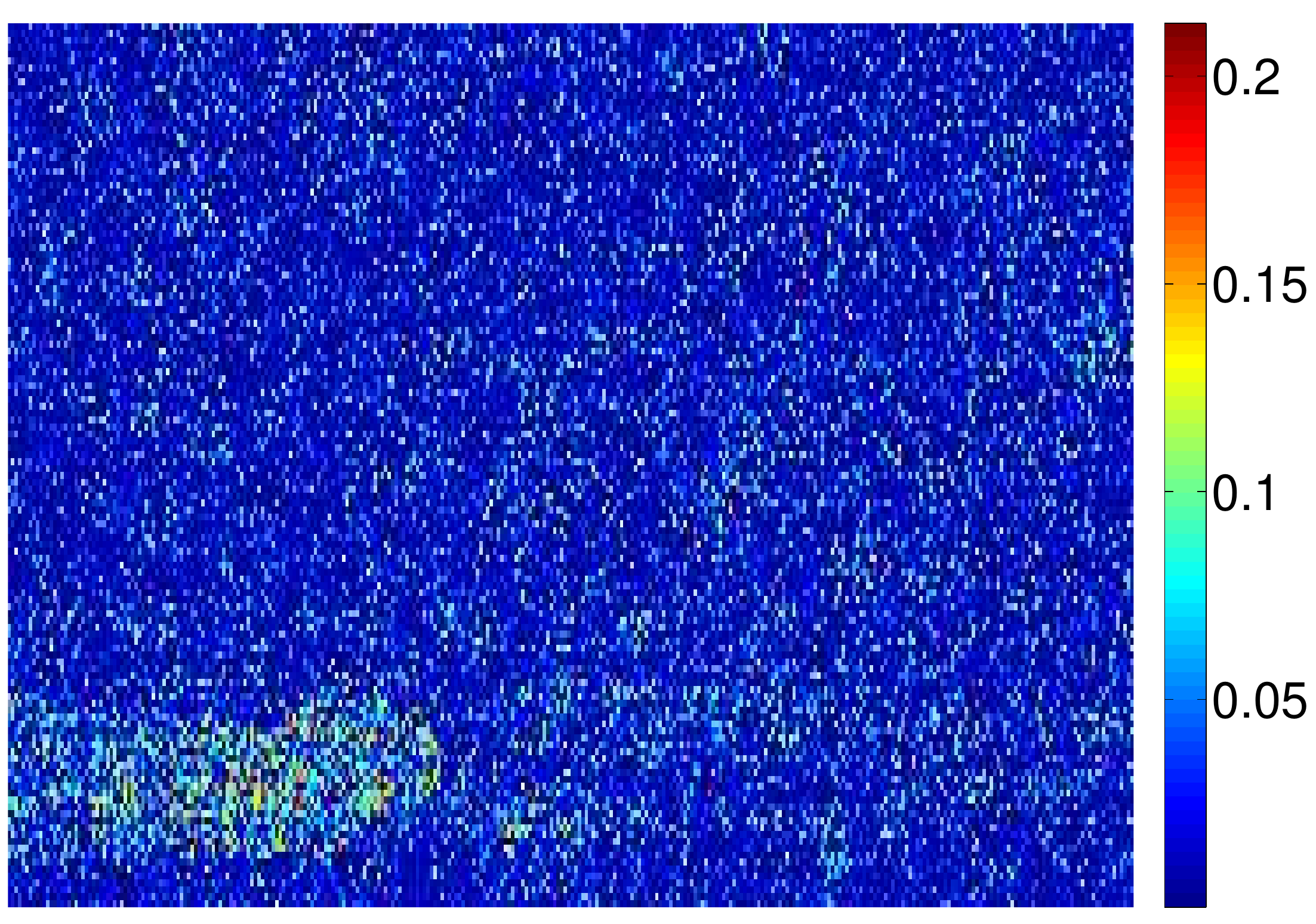}

\includegraphics[width=0.45\textwidth]{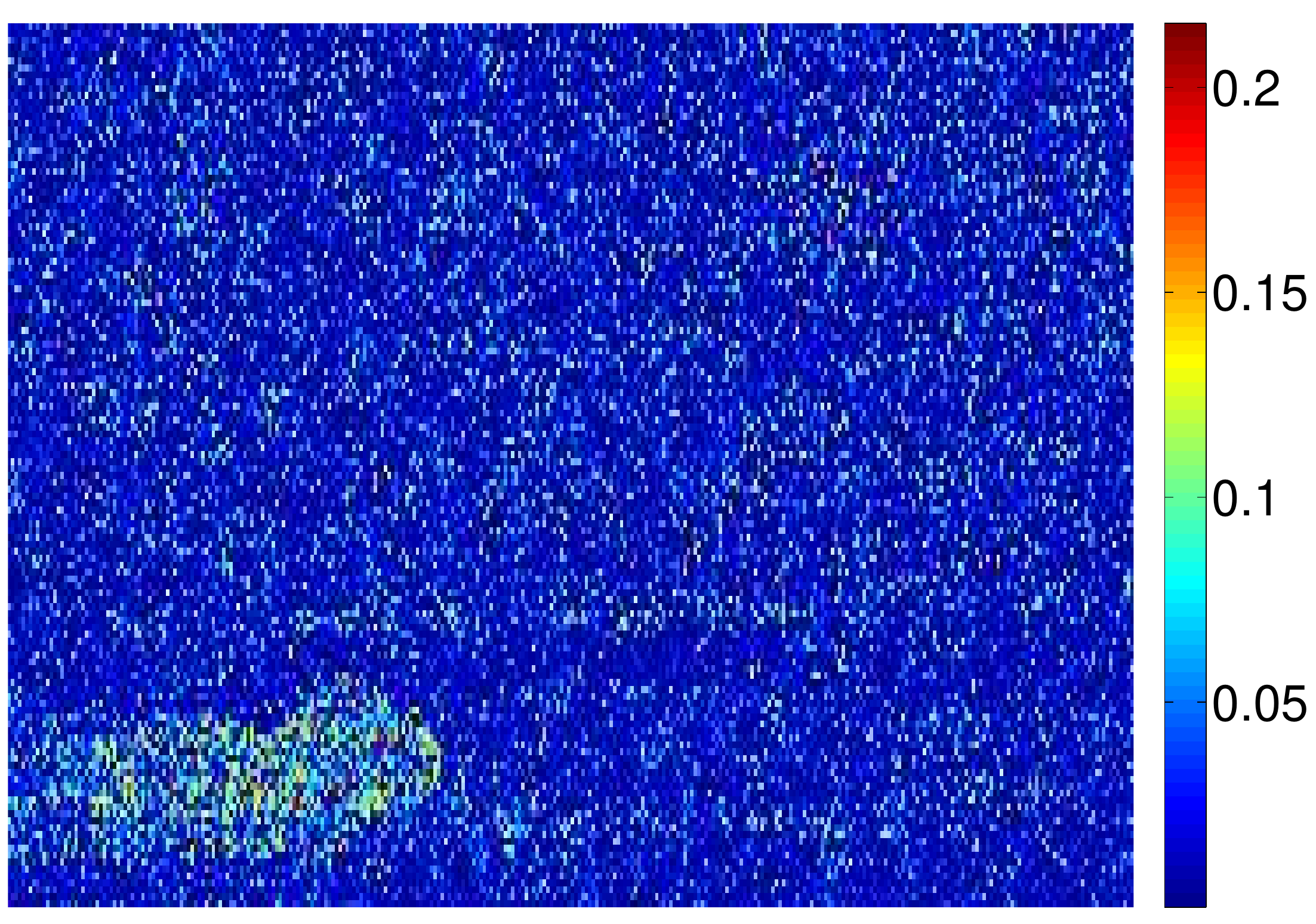}
\includegraphics[width=0.45\textwidth]{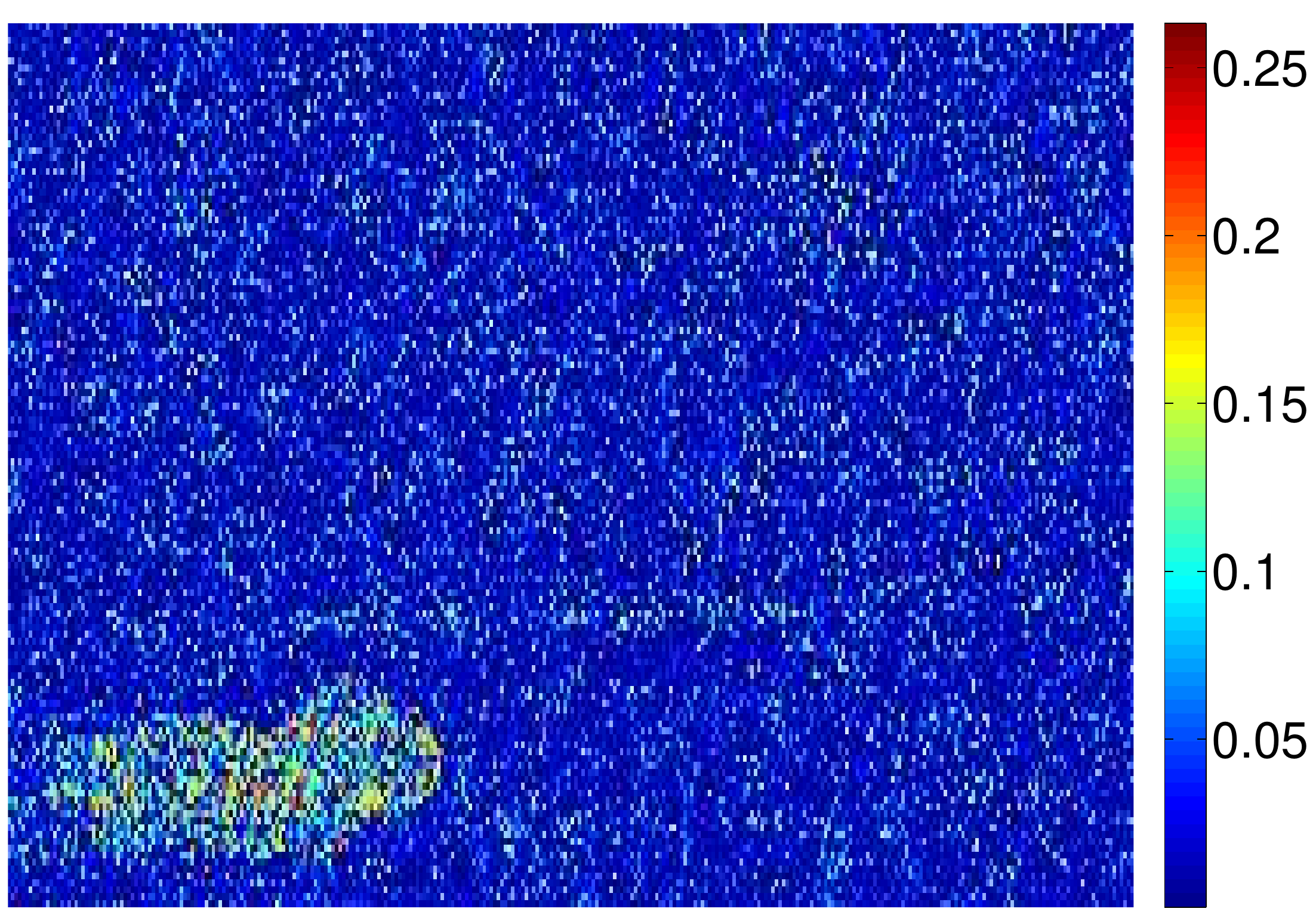}

\includegraphics[width=0.45\textwidth]{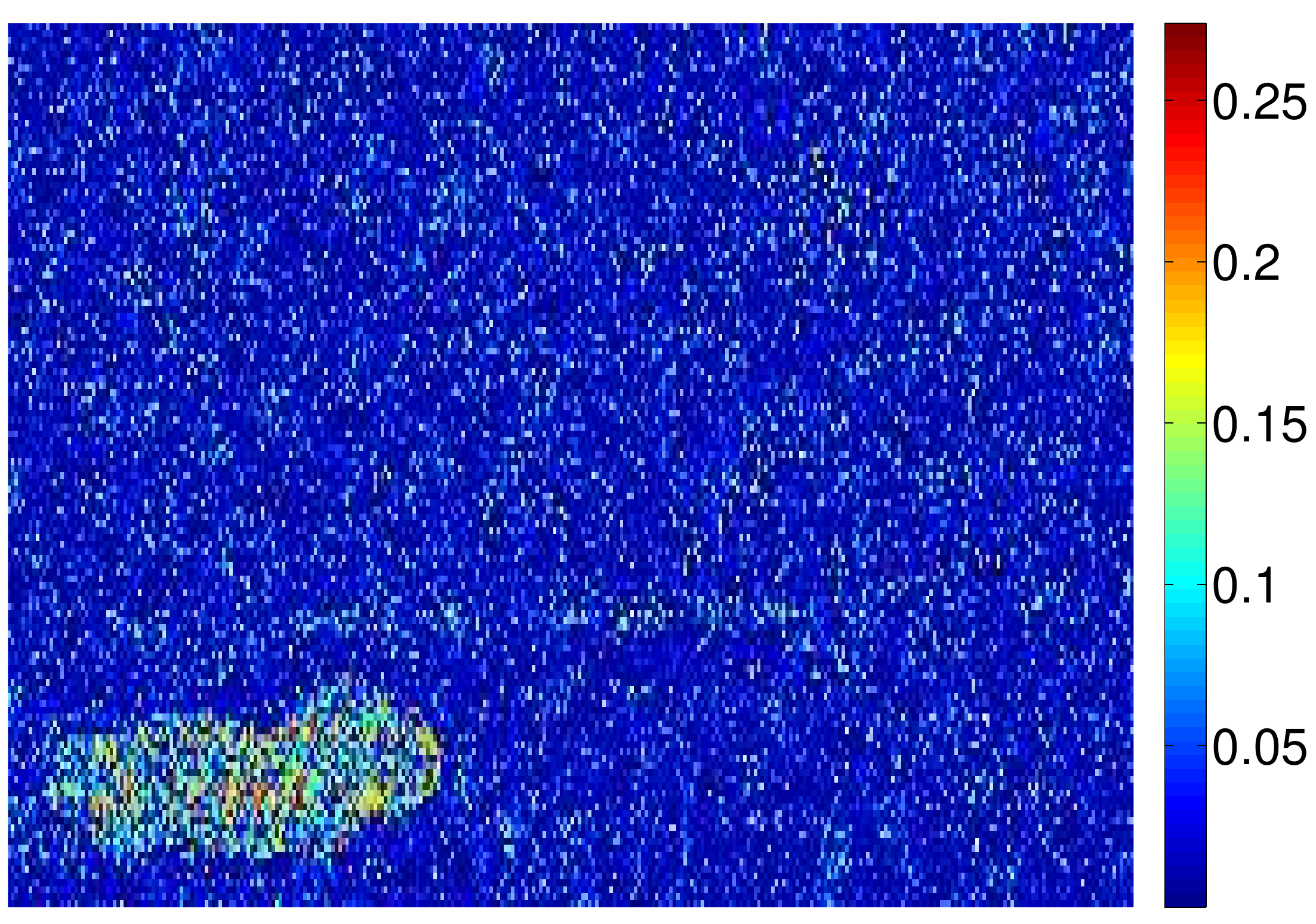}
\includegraphics[width=0.45\textwidth]{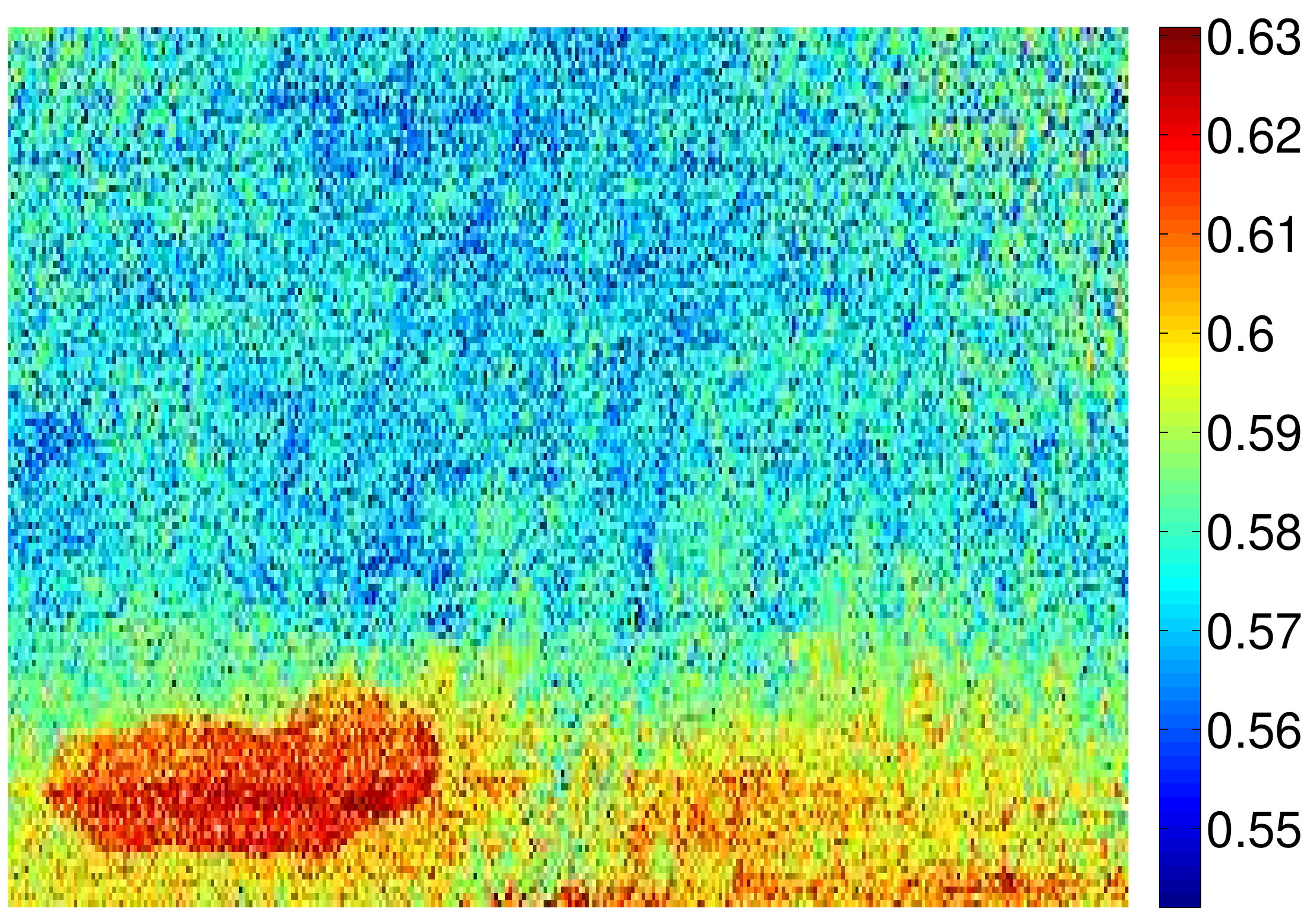}
\caption{From top to bottom and from left to right: the detections map of NMF-outlier, MNF, mixNMF, mixNMF-rs, mixNMF-rs2 and mixNMF-rs2-plsr. \label{fig:MIT_eg4_steps}}

\end{figure}


\subsection{Fabry-Perot Interferometer Sensor Data}
\begin{figure}[htbp!]

\centering
\includegraphics[width=0.4\textwidth]{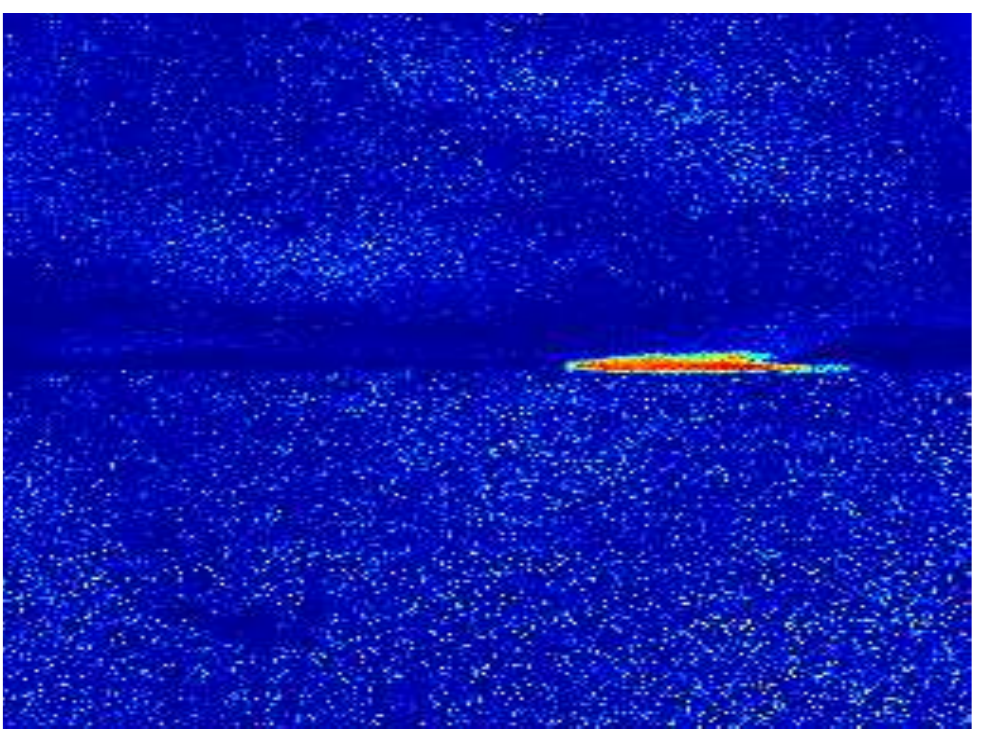}
\includegraphics[width=0.4\textwidth]{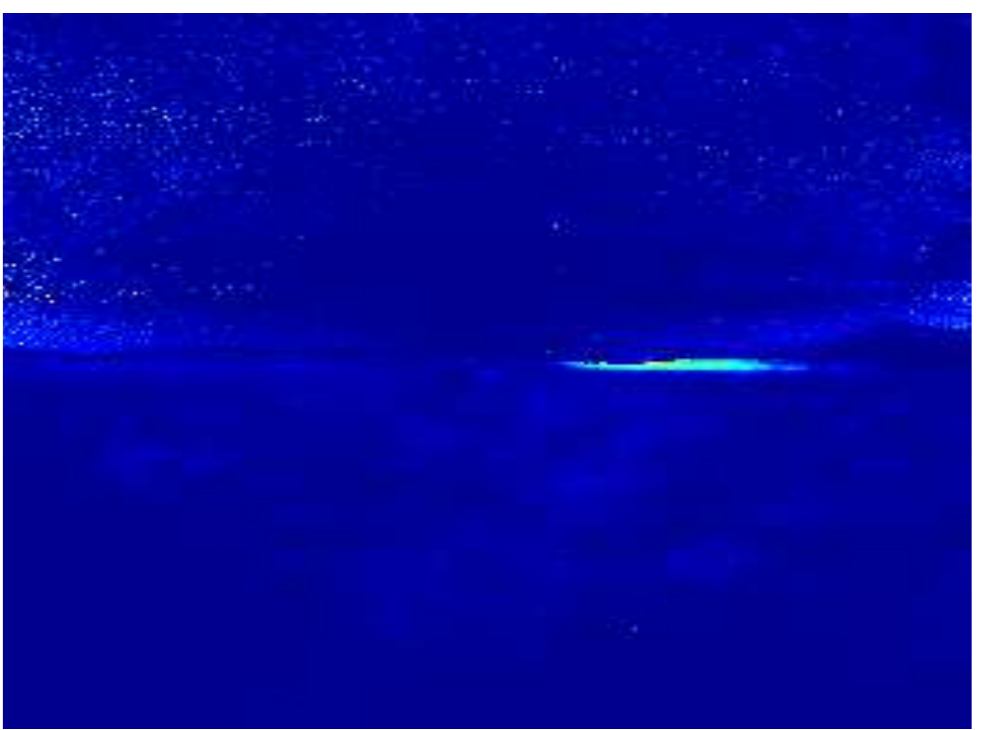}

\includegraphics[width=0.4\textwidth]{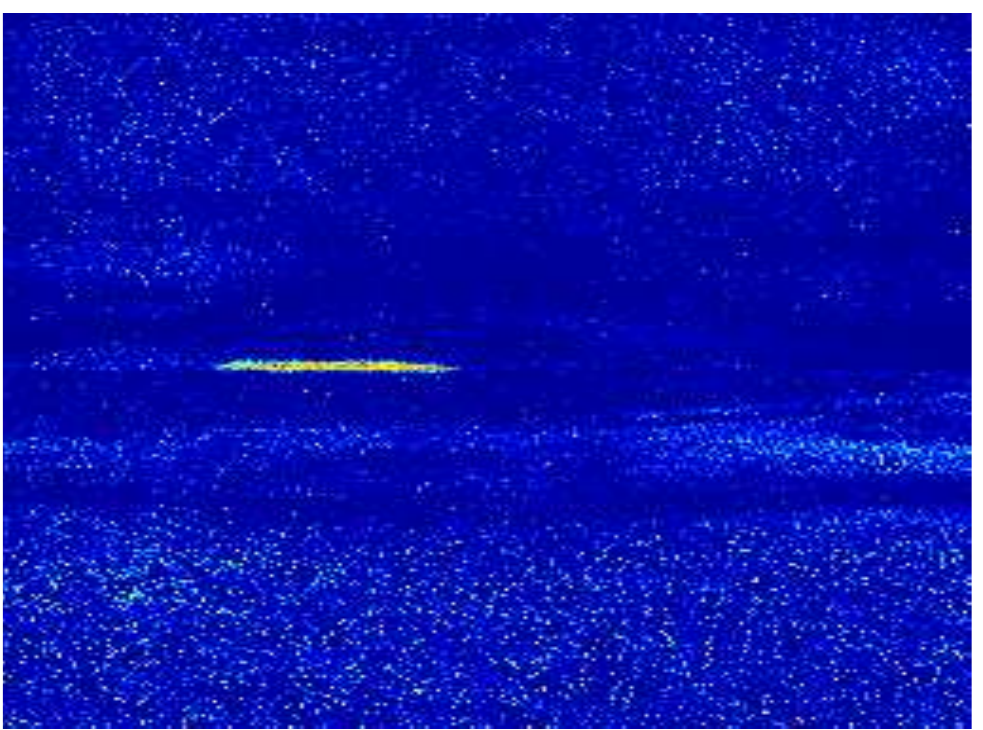}
\includegraphics[width=0.4\textwidth]{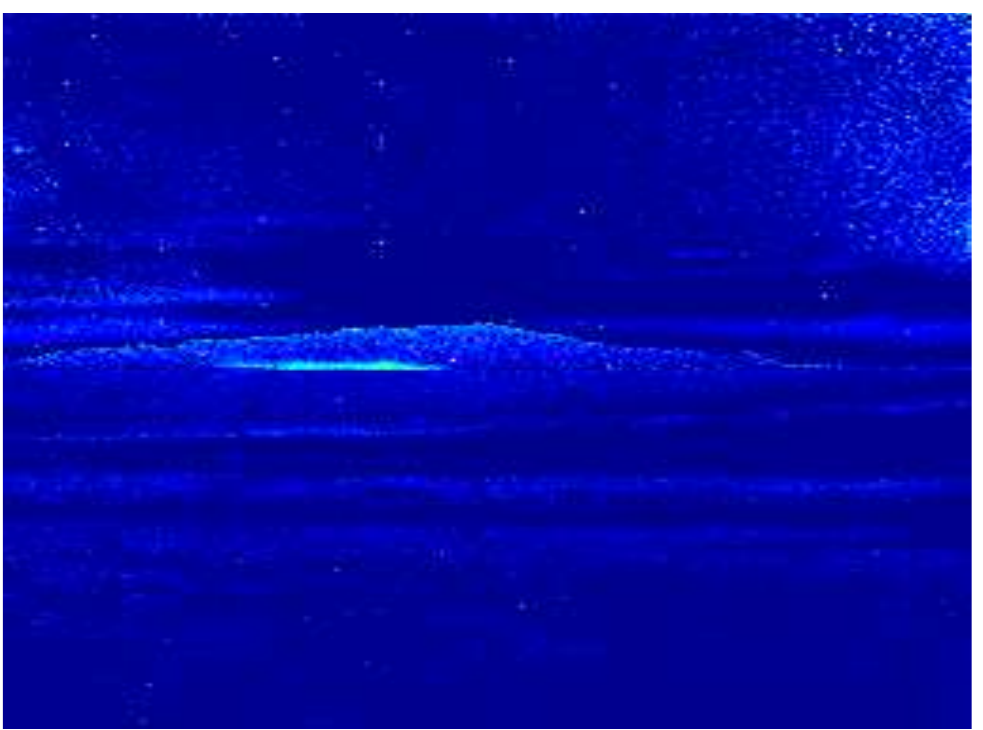}

\includegraphics[width=0.4\textwidth]{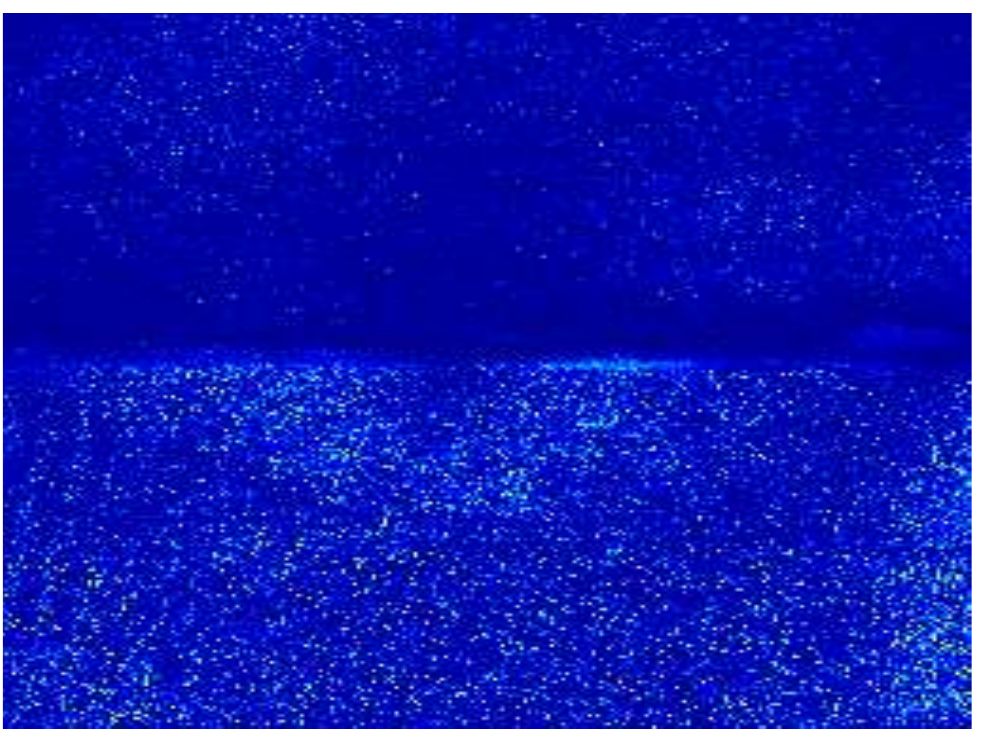}
\includegraphics[width=0.4\textwidth]{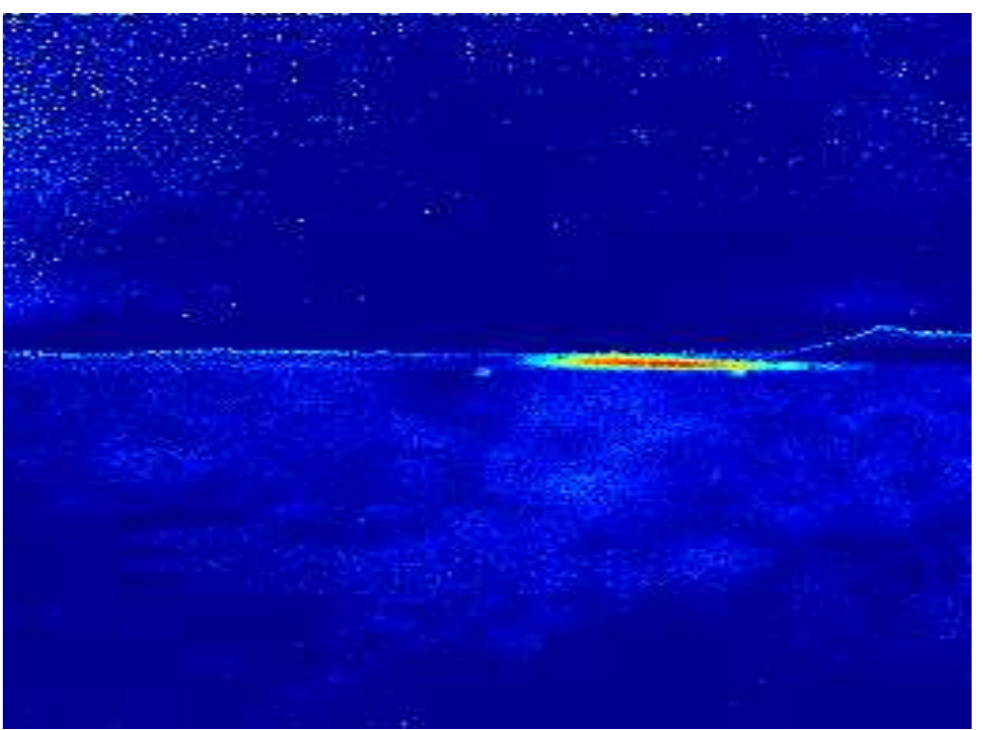}

\includegraphics[width=0.4\textwidth]{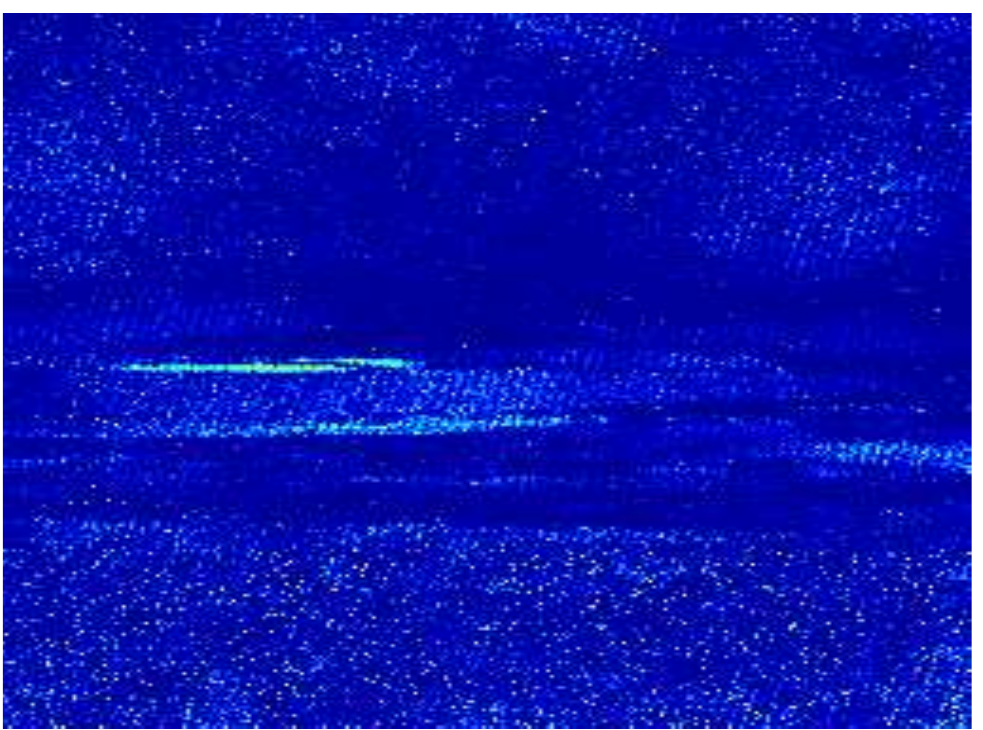}
\includegraphics[width=0.4\textwidth]{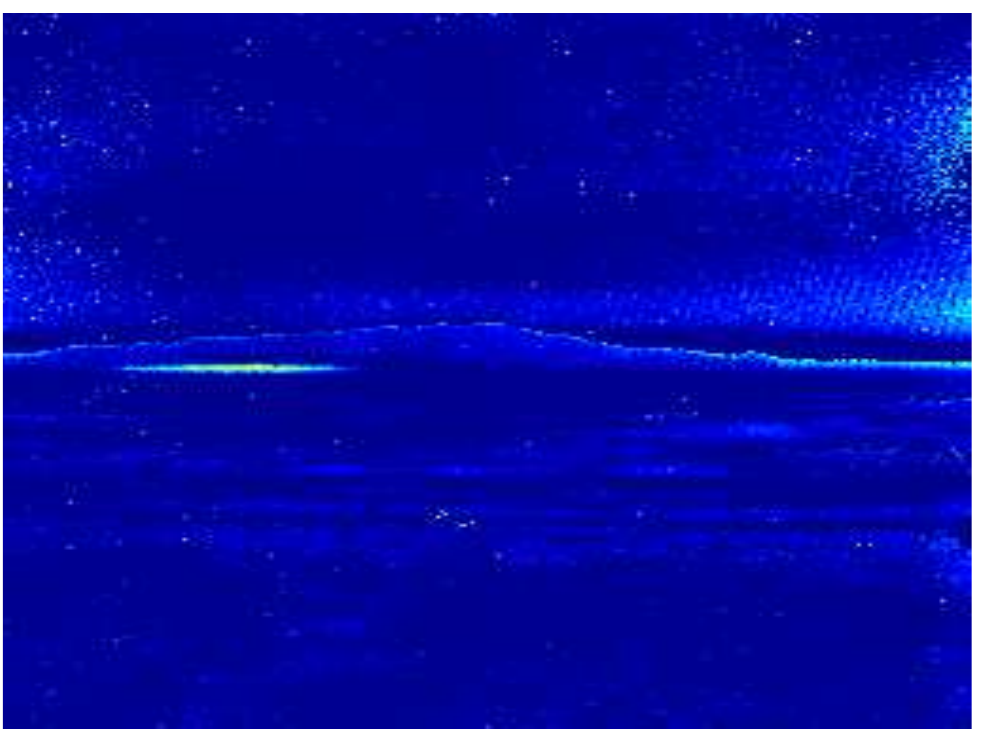}

\includegraphics[width=0.4\textwidth]{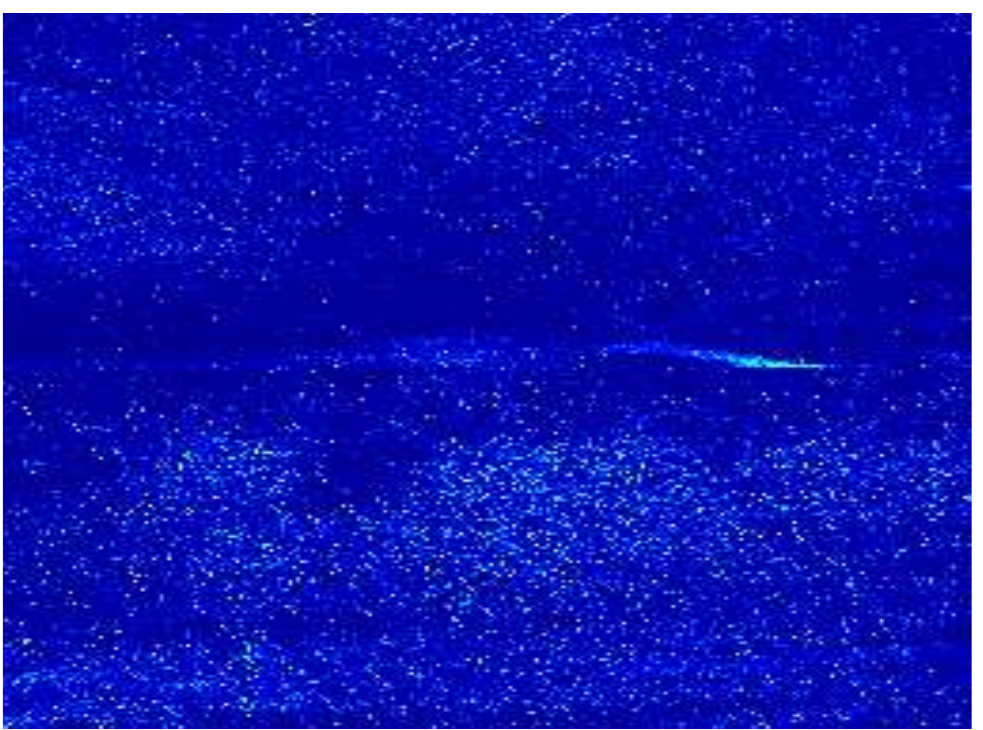}
\includegraphics[width=0.4\textwidth]{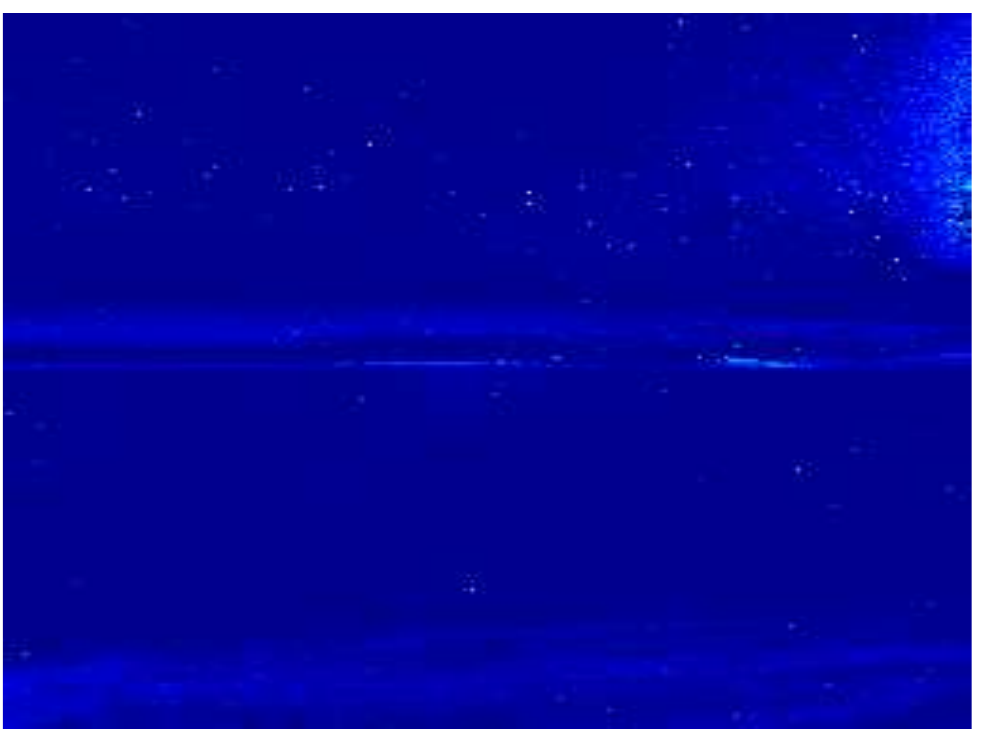}
\caption{Detection results by ACE (left column) and $T_{\text{mixLC}}$ (right column) on a fixed frame in each of the five hyperspectral movies of the Fabry-Perot data set. Note that (1) our detection map is consistently cleaner than ACE, and (2) our algorithm significantly outperformed ACE in the third movie.\label{results:FP}}

\end{figure}
\noindent{\underline{\em{Description}}}.
This dataset consists of five time series of radiance data, collected using an imaging Spectroradiometer that operates in the 8 -- 11 micron range under five different combinations of the kind of chemical material (TEP, or MeS, or GAA), the release amount (75 kg/burst or 150), and the sensor used. To generate each sequence, a predetermined quantity of simulant is released into the troposphere by using explosives (i.e.~a burst release). Successive data cubes are collected, beginning before the event and terminating when it is deemed that additional data collected will not augment scientific value. As a result, each hyperspectral time series contains several hundred frames of hyperspectral images, which all have $256\times 256$ spectra along 20 spectral bands.

\noindent{\underline{\em{Task}}. Detect and track the chemical plume.

\noindent{\underline{\em{Technique}}}.
We first check the time derivatives of the spectra (along the sequence) but did not observe anything useful (this indicates that the data is particularly challenging). We then applied Algorithm.~\ref{alg:cpd} with the mixLC estimator and 
compared our results with ACE~\cite{ACE}.

\noindent{\underline{\em{Results}}}.
The comparison is shown in Figure~\ref{results:FP}. It is obvious that the detection map by our algorithm is always much cleaner than that by ACE. Furthermore, in one case, our algorithm was able to clearly locate the plume while ACE could not (see Figure~\ref{results:FP}, third column).  In this experiment, we used the 5th frame for background modeling (via a union of lines) and the 35th frame for testing (for all five movies); future work will utilize multiple clean frames for joint background modeling.

\subsection{Anomaly Detection in Hyperspectral Movies}
\noindent{\underline{\em{Description}}}. We use hyperspectral movies available at CSR, collected by Johns Hopkins Applied Physics Lab, and made available through the NSF ATD program. These movies
are taken with an FTIR based long wave infrared sensor, recording a variety of releases of known chemicals in a gaseous, liquid and gaseous state. They have frames consisting of $128\times320\times120$ hyperspectral cubes, with one collected approximately every $8$ seconds.
The data consists of a desert scene where an unknown (to us) chemical is released at an unknown location at an unknown time (after the beginning of the movie), growing into a chemical plume.

\noindent{\underline{\em{Task}}.  Detect and track the chemical plume.

\noindent{\underline{\em{Technique}}}. We detect the spectra in the chemical plume as an anomaly with respect to an empirical model for the background. We use the first two frames of the movies (where we are told that no chemical plum is present) as our training set $X$, and use the techniques detailed in section \ref{sec:anomalydetection}  to construct $\hat\nu_{X}$, and detect anomalies.

\noindent{\underline{\em{Results}}}.
An example of anomaly detection is visualized in Figure \ref{fig:APL_anomaly}.
In the context of these movies, the intrinsic dimension of the data $d$ appears to be very low, typically $d\le 5$, as measured by Multiscale SVD \cite{MM:MultiscaleDimensionalityEstimationAAAI}-\cite{LMR:MGM1}.
Therefore the number of samples required in order to learn $\hat\nu_{X}$ is expected to be low, even if the high-dimensional space $R^p$ with $p=120$. In practice, with non-optimized Matlab code, the construction of $\hat\nu_{X}$, using the first frame as a training set (a much smaller set would be more than sufficient) takes about a minute, and the evaluation of $\hat\nu_{X}$ at all the spectra of a new frame takes a few seconds.
The anomaly detection requires the choice of a threshold, but not knowing the ground truth we cannot study ROC curves. However the choice of threshold did not seem to affect the results much, especially in the detection of the ``core'' part of the chemical plume, albeit it did affect the detection of (presumably less dense) regions of the plume: see Figure \ref{fig:APL_anomaly}, where we chose on purpose a particularly conservative threshold.
\begin{figure}[htbp!]
\centering
\includegraphics[height=0.4\textwidth]{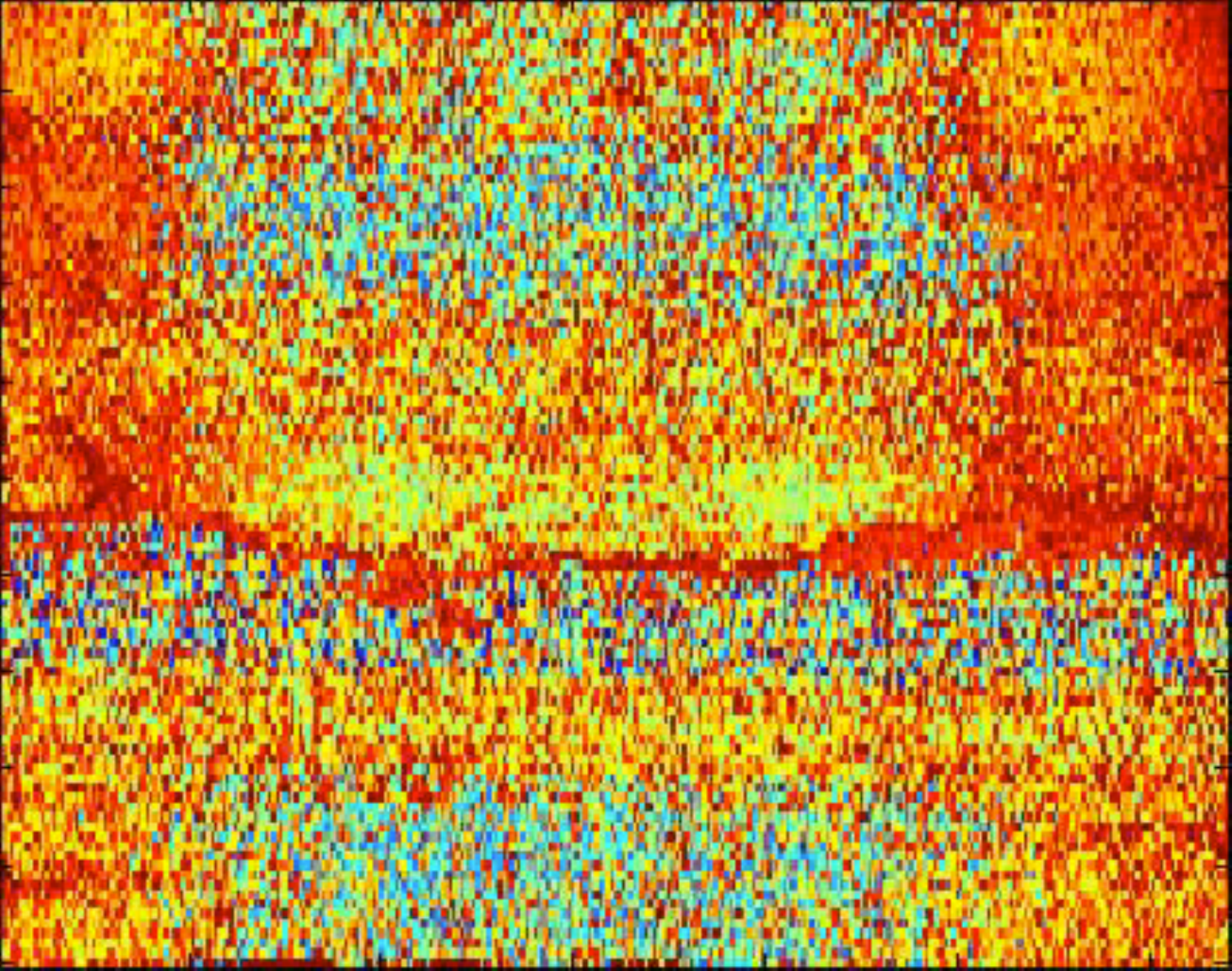}
\includegraphics[height=0.4\textwidth]{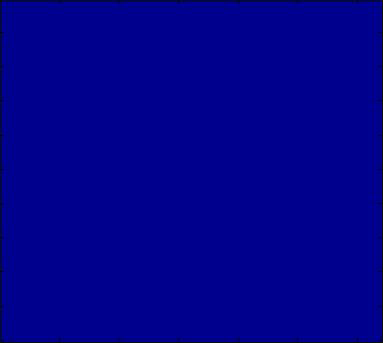}

\includegraphics[height=0.4\textwidth]{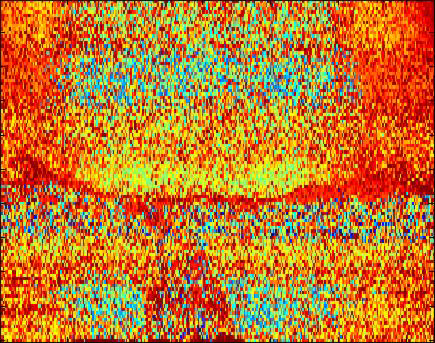}
\includegraphics[height=0.4\textwidth]{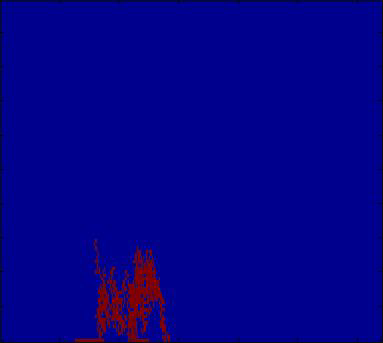}

\includegraphics[height=0.4\textwidth]{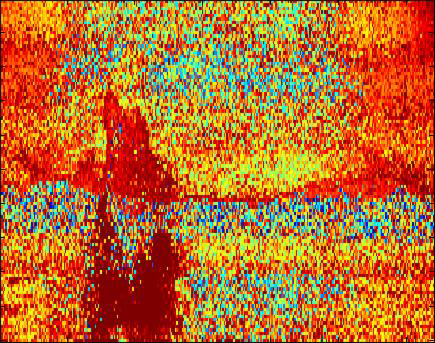}
\includegraphics[height=0.4\textwidth]{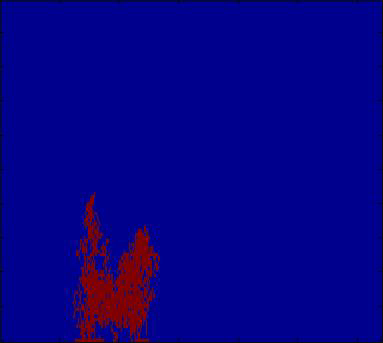}
\caption{Anomaly detection of a chemical plume of an unknown chemical appearing at an unknown time and location in a hyperspectral movie. No spatial registration among the frames is assumed. The chemical plume is detected as anomaly, evolving in time and space. Left column: log-likelihood score according to empirical GMRA-based model, with darker red meaning lower log-likelihood, i.e. higher probability of being an anomaly. Right column: thresholded version of the images on the left, with a non-optimized, conservative threshold, showing detection of the chemical plume. Ground truth was not provided for this data set. \label{fig:APL_anomaly}}
\vskip-0.7cm
\end{figure}

\section{Computational Complexity}
For the algorithms in section \ref{sec:mixmodels}, the algorithm~\cite{LBF} to estimate the subspaces modeling background has computational complexity is $O(c_1   m  n   (d  p + c_2 + \log(m  n)))$, where $d$ is the intrinsic dimension of the subspaces, $c_1$ and $c_2$ are two parameters (set to $10$ and $20$). Algorithm \cite{CM:CVPR2011} has similar cost and performance. Constructing the estimators $T_{\text{mixNMF}},\,T_{\text{mixNSS}}\,, T_{\text{mixLC}}$ requires $O(p^2),\,O((p+d)  d^2)$ and $O((p+d)  d^2)$ basic operations. Resampling has cost $O(m  n   \log (m   n))$, PLSR requires $O((p  m  n+l^2+p  l) l)$ operations, with $l$ the dimension chosen for PLSR (typically $O(1)$, independently of $p$).
The cost of the GMRA-based algorithms (see section \ref{sec:anomalydetection}) is: $O(C_d mnp\log(mn) d^2)$, where $C_d$ is a constant that depends exponentially in the intrinsic dimension $d$ of the subspaces, for constructing the GMRA; of $O(mn d^2)$ for constructing the estimator $\hat\mu_{\mathbf{s}}$ using low-rank Gaussians or KDE, and for evaluating it at new points; $O(mnp\log(mn))$ for computing the coefficients of new data needed to evaluate the likelihood.

In summary, all the algorithms we discuss run in time proportional to (up to logarithmic factors) the size $mnp$ of the hyperspectral data cube, with constants that depend on the intrinsic dimension $d$ of the data.

\section{Conclusion}
We have presented several ideas aimed at improving the current state-of-art in several tasks related to the analysis of hyperspectral images, in particular for background modeling, chemical plume detection and anomaly detection. We discussed the application of these algorithms to a variety of data sets, with state-of-art or better results (when ground truth was available). The proposed techniques are diverse, but are mostly motivated by the observation that hyperspectral data is often noisy but intrinsically low-dimensional, allowing one to use ideas from the dimension reduction and manifold learning algorithms originally considered in view of machine learning applications. In particular we use techniques for approximating data by mixtures of distributions on low-dimensional subspaces~\cite{SCC,LBF,CM:CVPR2011}, first with a small, fixed number of subspaces for background modeling, and then with more complex, multiscale mixtures of subspaces using GMRA~\cite{GMRA_ACHA12} and its extensions to the estimate of probability measures in high-dimensions.

\section*{Acknowledgment}
The authors thank Dr. Dimitris G.~Manolakis for sharing with them a manuscript~\cite{Manolakis-review-IRLW} and for his correspondence answering questions.

\bibliographystyle{plain}

\end{document}